\documentclass{article}

\usepackage[final]{neurips_2021}
\usepackage{todonotes}

\usepackage[utf8]{inputenc} 
\usepackage[T1]{fontenc}    
\usepackage{hyperref}       
\usepackage{url}            
\usepackage{booktabs}       
\usepackage{amsfonts}       
\usepackage{nicefrac}       
\usepackage{microtype}      
\usepackage{xcolor}         
\usepackage{caption}
\usepackage{subcaption}
\usepackage[utf8]{inputenc}
\usepackage[english]{babel}

\usepackage{amsthm}

\usepackage{tikz}
\usetikzlibrary{bayesnet}
\usetikzlibrary{arrows}
\usepackage{color}
\definecolor{darkgreen}{RGB}{44,160,44}

\definecolor{darkred}{RGB}{214,39,40}
\definecolor{darkorange}{RGB}{255,127,14}

\usetikzlibrary{backgrounds}

\usepackage{caption}
\usepackage{subcaption}

\usepackage{amsmath,amsfonts,bm}

\def\eqref#1{equation~\ref{#1}}

\def\1{\bm{1}}

\def\ra{{\textnormal{a}}}
\def\rb{{\textnormal{b}}}
\def\rc{{\textnormal{c}}}
\def\rd{{\textnormal{d}}}
\def\re{{\textnormal{e}}}

\def\rt{{\textnormal{t}}}

\def\rx{{\textnormal{x}}}
\def\ry{{\textnormal{y}}}
\def\rz{{\textnormal{z}}}

\def\rve{{\mathbf{e}}}

\def\rvx{{\mathbf{x}}}
\def\rvy{{\mathbf{y}}}
\def\rvz{{\mathbf{z}}}

\def\ve{{\bm{e}}}

\def\vx{{\bm{x}}}
\def\vy{{\bm{y}}}
\def\vz{{\bm{z}}}

\def\mW{{\bm{W}}}

\DeclareMathAlphabet{\mathsfit}{\encodingdefault}{\sfdefault}{m}{sl}
\SetMathAlphabet{\mathsfit}{bold}{\encodingdefault}{\sfdefault}{bx}{n}

\newcommand{\E}{\mathbb{E}}

\newcommand{\KL}{D_{\mathrm{KL}}}

\newcommand{\JSD}{D_{\mathrm{JSD}}}

\newtheorem{prop}{Proposition}

\newcommand{\remove}[1]{}

\title{An Information-theoretic Approach to Distribution Shifts}

\author{%
  Marco Federici\\
  AMLab\\
  University of Amsterdam\\
  \texttt{m.federici@uva.nl} \\
   \And
   Ryota Tomioka \\
   Microsoft Research\\
   Cambridge, UK\\
   \texttt{ryoto@microsoft.com} \\
   \AND
   Patrick Forr\'e \\
   AI4Science Lab, AMLab \\
   University of Amsterdam \\
   \texttt{p.d.forre@uva.nl} \\

}

\begin{document}

\maketitle

\begin{abstract}
Safely deploying machine learning models to the real world is often a challenging process. 
Models trained with data obtained from a specific geographic location tend to fail when queried with data obtained elsewhere, agents trained in a simulation can struggle to adapt when deployed in the real world or novel environments, and neural networks that are fit to a subset of the population might carry some selection bias into their decision process.
In this work, we describe the problem of data shift from a novel information-theoretic perspective by (i) identifying and describing the different sources of error, (ii) comparing some of the most promising objectives explored in the recent domain generalization and fair classification literature. From our theoretical analysis and empirical evaluation, we conclude that the model selection procedure needs to be guided by careful considerations regarding the observed data, the factors used for correction, and the structure of the data-generating process.
\end{abstract}

\section{Introduction}
\label{intro}

One of the most common assumptions for machine learning models is that the training and test data are independently and identically sampled (IID) from the same distribution. In practice, this assumption does not hold in many practical scenarios \citep{Bengio2019}.
A machine learning model trained to recognize land usage from satellite images using pictures from the early 2000s may struggle to recognize the style of modern architectures \citep{Christie2017}, data collected on a limited set of hospitals may not be representative of the variation introduced by the use of different machines or procedures \citep{Zech2018, Beede2020}.
Other kinds of distribution shifts are more subtle and difficult to recognize despite having a noticeable impact on the model's predictive performance.
Examples include under-represented or over-represented population groups \citep{Popejoy2016, Buolamwini2018} or biased annotations collected from crowd-sourcing services \citep{Zhang2017, Xia2020}.

Different approaches in literature address these issues by using some external source of knowledge such as domain or environment annotations \citep{Wang2018}, protected attributes \citep{Mehrabi2019} or sub-population groups \citep{Santurkar2020} to reduce bias and minimize the model error outside of the training distribution. Despite the progress in the field of domain generalization literature, \citet{Gulrajani2020} has shown that the effectiveness of some of the most common algorithms heavily relies on the hyper-parameter tuning strategy, revealing limitations of examined models when compared to a more traditional empirical risk minimization strategy. 
One of the major issues behind this observed behavior is a lack of clarity and applicability for the underlying assumptions regarding the problem statement and the data-generating process \citep{Zhao2019, Rosenfeld2020, Mahajan2020}.

With this work, we aim to present a new perspective on the problem to analyze and clarify the fundamental differences between some of the most common approaches by:
\begin{enumerate}
    \item Introducing a novel information-theoretical framework to describe the problem of distribution shift and connecting it to the test error and its components (section~\ref{sec:framework}).
    \item Analyzing four main families of objectives and describing some of their guarantees and assumptions (section~\ref{sec:criteria}).
    \item Demonstrating that the effectiveness of different criteria is determined by the structure of the underlying data-generating process (section~\ref{sec:experiments}).
    \item Showing that the results obtained by popular models designed according to the aforementioned criteria can drastically differ from the theoretically expected performance (section~\ref{sec:experiments}).
\end{enumerate}

The analysis of the \textit{bottleneck}, \textit{independence}, \textit{sufficiency}, and \textit{separation} criteria reveals that some of the most popular models can systematically fail to reduce the test error even for simple datasets. No unique objective is simultaneously optimal for every problem, but additional knowledge about the selection procedure and better approximations can help to mitigate the bias.

\subsection{Problem Statement}

Consider $\rvx$ and $\rvy$ as the features and targets respectively with joint density $p(\rvx,\rvy)$ for the predictive problem of interest. 
Let $\rt$ be a binary random variable representing which data is selected for training ($\rt=1$) and which is not ($\rt=0$). We will refer to $p(\rvx,\rvy|\rt=1)$ as the joint \textit{Training distribution} and $p(\rvx,\rvy|\rt=0)$ as the \textit{Test distribution} that are induced by the selection $\rt$.
Throughout the paper, we will consider selections $\rt$ that can be expressed as a function of the features $\rvx$, targets $\rvy$, independent noise $\epsilon$ and other variables $\rve$, which represent other factors that are affecting the data collection procedure (e.g. geographic location, time intervals, population groups). 

Let $q(\rvy|\rvx)$ represent a learnable model for the predictive distribution $p(\rvy|\rvx)$. We use the Kullback-Leibler divergence to express the \textit{Train} 
and \textit{Test} 
error respectively\footnote{Further details regarding the convention used for conditional KL-divergence can be found in appendix~\ref{app:notation}}:
\begin{align}
    \underbrace{\KL(p^{\rt=1}_{\rvy|\rvx}||q_{\rvy|\rvx})}_{\text{Train error}} :=\KL(p(\rvy|\rvx,\rt=1)||q(\rvy|\rvx)),\\
    \underbrace{\KL(p^{\rt=0}_{\rvy|\rvx}||q_{\rvy|\rvx})}_{\text{Test error}}:=\KL(p(\rvy|\rvx,\rt=0)||q(\rvy|\rvx)).
\end{align}
Although only the train distribution can be accessed at training time, we are interested in learning models $q(\rvy|\rvx)$ that result in small test error.

\subsection{Characterizing Distribution Shift}
As a first step, we characterize the difference between training and test distribution as a function of the selection.
The effect that the selection has in the joint distribution $p(\rvx,\rvy)$ can be quantified by considering the mutual information between the features-target pair $\rvx\rvy$ and the selection variable $\rt$, which can be also expressed as a Kullback-Leibler divergence:
$
    I(\rvx\rvy;\rt) = \KL(p(\rvx,\rvy|\rt)|p(\rvx,\rvy))
$.
This measure of distribution shift intuitively represents how many bits the selection variable carries about the joint distribution of targets and features or, equivalently, how much the joint density $p(\rvx,\rvy)$ has changed as a result of the selection.

Using the chain rule of mutual information, one can express distribution shift 
as the sum of two separate components: 
\begin{align}
    \underbrace{I(\rvx\rvy;\rt)}_{\text{Distribution shift}} = \underbrace{I(\rvx;\rt)}_{\text{Covariate shift}} + \underbrace{I(\rvy;\rt|\rvx)}_{\text{Concept shift}},
\end{align}
which can be interpreted as the amount of \textit{covariate shift} \citep{Shimodaira2000} and \textit{concept shift}  \citep{Widmer1996, Moreno2012}, respectively. These two quantities refer to the changes in the predictive distribution $p(\rvy|\rvx)$ and the marginal features distribution $p(\rvx)$ respectively, which add up to represent the changes in the joint distribution. 

Whenever the data selection is perfectly IID, the selection variable can be expressed as a function of some independent noise ($\rt=f(\epsilon)$) and the corresponding distribution shift $I(\rvx\rvy;\rt)$ is zero. On the other hand, if the data collection procedure has been influenced by other factors, we do not have such a guarantee, even when the selection does not depend on features and targets directly ($\rt=f(\rve,\epsilon)$).

\begin{figure}[!t]
\centering
    \begin{subfigure}{.5\textwidth}
        \centering
        \includegraphics[width=1.1\columnwidth]{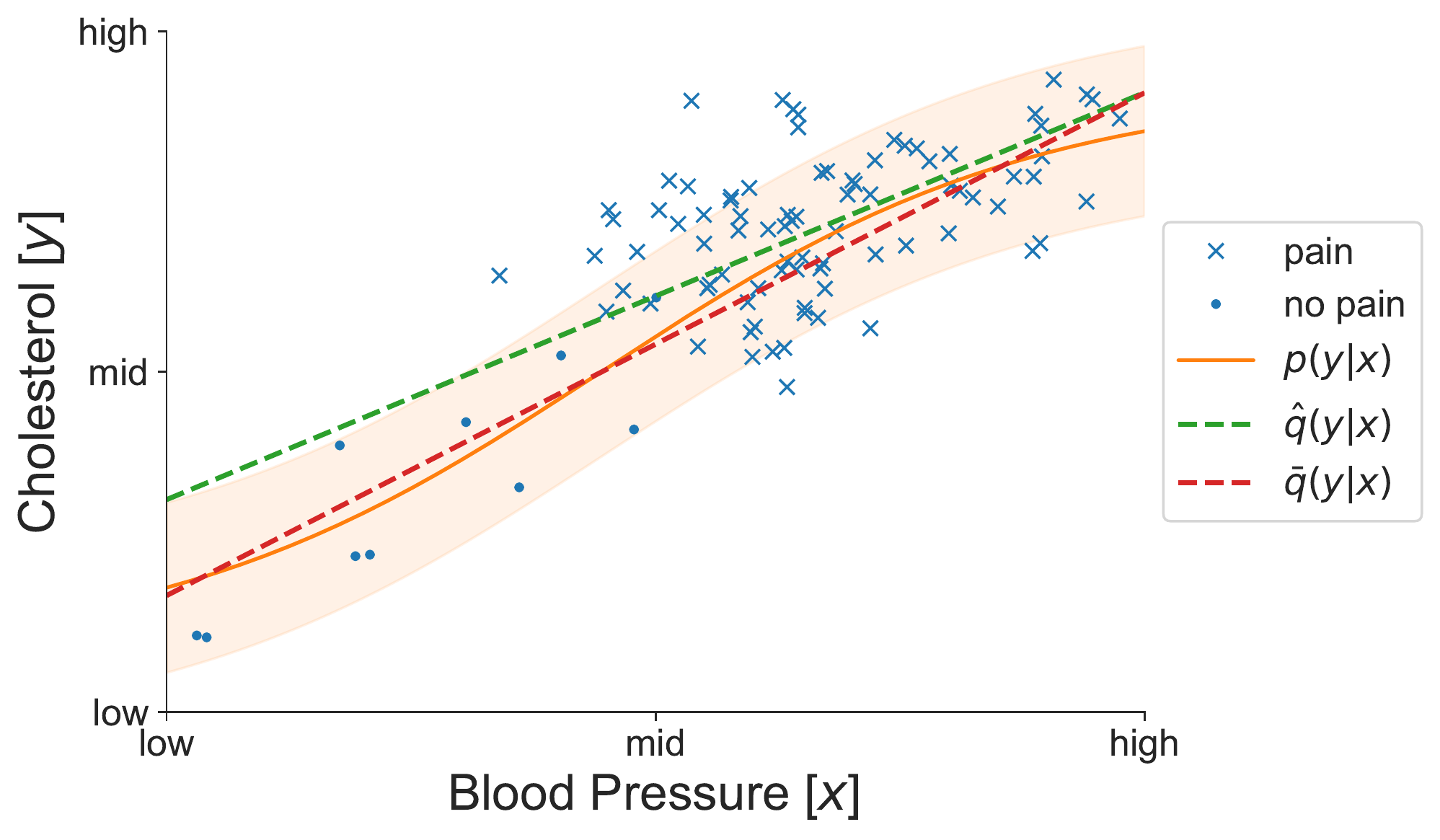}
        \vspace{-0.7cm}
        \caption{}
        \label{fig:example}
    \end{subfigure}%
    \begin{subfigure}{.5\textwidth}
        \centering
        \input{figures/graphical_model_1}
        \caption{}
        \label{fig:example_graph}
    \end{subfigure}%
    \caption{(a) Example of distribution shift due to selection bias (chest pain complaints) for the prediction of cholesterol levels (y-axis) given blood pressure (x-axis). A maximum likelihood approach (green dashed line) results in a model that over-estimates $\ry$ for any given $\rx$ when compared to the best model in the class (red dashed line). (b) Example of a data generating process in which the features $\rvx$ are composed of two parts $\rx_1$ and $\rx_2$. When both feature components $\rx_1$ and $\rx_2$ are observed, concept shift can be strictly positive. Discarding information about $\rx_2$ will reduce the effect of the selection bias, while removing $\rx_1$ might increase it $0=I(\rvy;\rt|\rx_1)\le I(\rvy;\rt|\rx_1\rx_2)\le I(\rvy;\rt|\rx_2)$. }
\end{figure}

\section{Information-theoretic Framework}  
\label{sec:framework}
Using the quantities described in the previous sections, we can show that the sum train and test error is lower-bounded by the amount concept shift:
\begin{prop}
    \label{prop:ood_error}
    For any model $q(\rvy|\rvx)$ and $\alpha:=\min\{p(\rt=0),p(\rt=1)\}$:
    \begin{align}
        \KL(p^{\rt=1}_{\rvy|\rvx}||q_{\rvy|\rvx}) + \KL(p^{\rt=0}_{\rvy|\rvx}||q_{\rvy|\rvx}) \ge \frac{1}{1-\alpha}I(\rvy;\rt|\rvx),
    \end{align}
\end{prop}

As a consequence, whenever the selection induces concept shift $I(\rvy;\rt|\rvx)>0$, any sufficiently flexible model $\hat q(\rvy|\rvx)$ trained using a maximum likelihood approach must incur in strictly positive test error.

Intuitively, whenever the selection induces a change in the predictive distribution ($p(\rvy|\rvx)\neq p(\rvy|\rvx,\rt=1)$) fitting the model to the training distribution will incorporate the selection bias into the model prediction, necessarily resulting in errors when evaluated on the test distribution, as shown in  the example reported in figure~\ref{fig:example}. 

The different approaches analyzed in this work are based in the introduction of a \textit{latent representation} $\rvz$, which allows for the definition of different regularization strategies.
Using the latent representation as an intermediate variable, the model $q(\rvy|\rvx)$ can be re-parametrized with an \textit{encoder} $q(\rvz|\rvx)$ and a \textit{latent classifier}\footnote{The name classifier will be used for convenience to express both classification and regression problems.} $q(\rvy|\rvz)$:
\begin{align}
    q(\rvy|\rvx) = \E_{\vz\sim q(\rvz|\rvx)}\left[q(\rvy|\rvz=\vz)\right].
\end{align}
The effect of the introduction of a latent representation can be observed by expressing the training and test error as a function of the encoder and the classifier\footnote{The expressions in equations~\ref{eq:latent_train} and \ref{eq:latent_ood} hold with equality for deterministic encoders as shown in appendix~\ref{app:proofs}.}.
\begin{prop}
    For any encoder $q(\rvz|\rvx)$ and classifier $q(\rvy|\rvz)$:
\begin{align}
   \KL(p^{\rt=1}_{\rvy|\rvx}||q_{\rvy|\rvx})\le \underbrace{I_{\rt=1}(\rvx;\rvy|\rvz)}_{\text{Train information loss}} + \underbrace{\KL(p_{\rvy|\rvz}^{\rt=1}||q_{\rvy|\rvz})}_{\text{Latent train error}}\label{eq:latent_train}\\
     \KL(p^{\rt=0}_{\rvy|\rvx}||q_{\rvy|\rvx})\le \underbrace{I_{\rt=0}(\rvx;\rvy|\rvz)}_{\text{Test information loss}} + \underbrace{\KL(p_{\rvy|\rvz}^{\rt=0}||q_{\rvy|\rvz})}_{\text{Latent test error}}.\label{eq:latent_ood}
\end{align}\label{prop:decomposition}\end{prop}
\vspace{-0.2cm}
The two terms $I_{t=1}(\rvx;\rvy|\rvz)$  and $I_{t=0}(\rvx;\rvy|\rvz)$ represent the amount of predictive information that is lost by encoding the features $\rvx$ into $\rvz$ on train and test distribution respectively, while $\KL(p_{\rvy|\rvz}^{\rt=1}||q_{\rvy|\rvz})$ and $\KL(p_{\rvy|\rvz}^{\rt=0}||q_{\rvy|\rvz})$ refer to the train and test error when using $\rvz$ instead of the original observations as the predictive features, which will be referred to as \textit{latent training error} and \textit{latent test error} respectively.
Test information loss and latent test error capture two intrinsically different kinds of error. The former indicates  the increase in the prediction uncertainty as a result of the encoding procedure, while the latter represents the discrepancy between the model $q(\rvy|\rvz)$ and the latent test predictive distribution $p(\rvy|\rvz,\rt=0)$.

\subsection{Latent test error}
Minimizing the latent test error makes the latent predictor $q(\rvy|\rvz)$ approach the test latent predictive distribution $p(\rvy|\rvz, \rt=0)$.
We can show that the Jensen-Shannon divergence between the two distributions is upper-bounded by a monotonic function of the latent training error and the amount of concept shift in the latent space (\textit{latent concept shift} $I(\rvy;\rt|\rvz)$):
\begin{prop}
    \label{prop:uebound}
    For any $q(\rvz|\rvx)$, $q(\rvy|\rvz)$ and any representation $\vz$ that satisfies $p(\rvz=\vz|\rt=0)>0$ and $p(\rvz=\vz|\rt=1)>0$:
    \begin{align}
        \left(\sqrt{\frac{1}{2\alpha}I(\rvy;\rt|\rvz=\vz)} +\sqrt{\frac{1}{2}\KL(p_{\rvy|\rvz=\vz}^{\rt=1}||q_{\rvy|\rvz=\vz})}\right)^{2} \ge \JSD(p^{\rt=0}_{\rvy|\rvz=\vz}||q_{\rvy|\rvz=\vz}).
    \end{align}
\end{prop}
Whenever the Jensen-Shannon divergence between the test predictive distribution $p(\rvy|\rvz,\rt=0)$ and the classifier $q(\rvy|\rvz)$ is small, the latent test error (measured in terms of KL-divergence) must also be small at least for the regions that have positive probability according to both train $p(\rvx|\rt=1)$ and test $p(\rvx|\rt=0)$ data distributions. Since the train predictive distribution $p(\rvy|\rvx,\rt=1)$ is not defined for $\vx$ that have zero probability on the train distribution, we have no guarantees regarding the model predictions in those regions unless other inductive biases are considered.

The left hand side of the expression in proposition~\ref{prop:uebound} can be minimized by addressing $I(\rvy;\rt|\rvz)$ and $\KL(p_{\rvy|\rvz}^{\rt=1}||q_{\rvy|\rvz})$ with respect to the encoder and classifier respectively. In other words, we can minimize the latent test error by simultaneously fitting $q(\rvy|\rvz)$ to the train predictive distribution $p(\rvy|\rvz,\rt=1)$ and minimizing the latent concept shift induced by the encoder $q(\rvz|\rvx)$.
When the latent concept shift and latent test error approach zero, the latent train predictive distribution $p(\rvy|\rvz,\rt=1)$ approaches the true (unselected) predictive distribution $p(\rvy|\rvz)$, and so does the modeled classifier $q(\rvy|\rvz)$.
In this ideal scenario, the only source of test error is the additional uncertainty that is due to the information lost in the encoding procedure.

\subsection{Minimizing the information loss}

Since losing information generally results in increased test error (equation~\ref{eq:latent_ood}), we will consider objectives that discard the minimal amount of information required to reduce the latent concept shift. This can be done by minimizing the KL-divergence between the training predictive distribution $p(\rvy|\rvx;\rt=1)$ and the latent classifier model $q(\rvy|\rvz)$:
\begin{align}
    \min_{q(\rvz|\rvx)} I_{\rt=1}(\rvx;\rvy|\rvz) &=\min_{q(\rvz|\rvx),q(\rvy|\rvz)} \KL(p_{\rvy|\rvx}^{\rt=1}||q_{\rvy|\rvz})-KL(p_{\rvy|\rvz}^{\rt=1}||q_{\rvy|\rvz}) \nonumber\\
    &\le \min_{q(\rvz|\rvx),q(\rvy|\rvz)} \KL(p_{\rvy|\rvx}^{\rt=1}||q_{\rvy|\rvz}). 
    \label{eq:bounds_info}
\end{align}
In addition to reducing the amount of information lost in the encoding procedure, minimizing the right hand side of the expression in equation~\ref{eq:bounds_info} makes the model $q(\rvy|\rvz)$ approach the latent predictive distribution $p(\rvy|\rvz,\rt=1)$.
Note that only the selected training distribution $p(\rvy,\rvx| \rt=1)$ is available at training time. In practice, minimizing the train information loss also generally results in decreased test information loss since the same features are usually informative for both train and test distributions.

\subsection{A General Loss function}
To summarize, the overall objective consists in finding the maximally informative representation that minimizes latent concept shift so that the same learned predictor $q(\rvy|\rvz)$ can perform similarly on both train and test settings.
This is achieved by (i) minimizing the amount of latent concept shift induced by the representation, (ii) maximizing the amount of predictive information in the representation, and (iii) matching $q(\rvy|\rvz)$ and the training latent predictive distribution $p(\rvy|\rvz, \rt=1)$.
The three requirements can be enforced by considering a loss function with the following form:
\begin{align}
    \mathcal{L}(q_{\rvz|\rvx},q_{\rvy|\rvz};\lambda) = \KL(p_{\rvy|\rvx}^{\rt=1}||q_{\rvy|\rvz})+\lambda \mathcal{R}(q_{\rvz|\rvx}),
    \label{eq:general_obj}
\end{align}
in which, the first term addresses (ii) and (iii), while the second term represent a regularization term that acts on the encoder $q(\rvz|\rvx)$ to minimize latent concept shift $I(\rvy;\rt|\rvz)$, following (i). For a sufficiently flexible family of latent predictors, requirement (iii) depends only $q(\rvy|\rvz)$, while the hyper-parameter $\lambda$ defines the trade-off between latent concept shift (i) and predictive information loss (ii).

\section{Regularization Criteria}
\label{sec:criteria}

Since only selected data ($\rt=1$) is accessible at training time, the latent concept shift can not be computed or minimized directly. Most of the approaches considered in this analysis make use of a regularization $\mathcal{R}(q_{\rvz|\rvx})$ that is based on the observation of an additional variable $\rve$ which relates to the selection criteria. This variable is usually referred to as \textit{domain} or \textit{environment}\footnote{In contrast with the domain adaptation and generalization literature, we will consider the more general case in which $\rve$ is represented by a vector.} in the domain adaptation and generalization literature, while the name \textit{protected attribute} is used in the context of fair classification.
We will refer to this variable $\rve$ as \textit{environmental factors} in the following sections.

We analyze four families of criteria proposed in the representation learning \citep{Tishby2015}, domain generalization \citep{Koyama2020} and fair classification \citep{Barocas2018} literature focusing on their underlying assumptions and theoretical guarantees.
The different regularization strategies and models can be seen as specific instance of the loss function in equation~\ref{eq:general_obj}. 
An empirical comparison between instances of the different criteria can be found in section~\ref{sec:experiments}, proofs are reported in appendix~\ref{app:proofs}, while the relation between the reported criteria is further discussed in appendix~\ref{app:crit_relations}.

\subsection{Information Bottleneck Criterion}
Combining the results from propositions~\ref{prop:ood_error} and \ref{prop:decomposition} we can infer that reducing the latent test error necessarily requires the representation to discard some train predictive information ($I_{\rt=1}(\rvx;\rvy|\rvz)$>0). This is because a lossless encoder $\hat{q}(\rvz|\rvx)$ together with an optimal latent classifier $\hat{q}(\rvy|\rvz)$ would result in an overall model $\hat{q}(\rvy|\rvx)$ that matches the train distribution $p(\rvy|\rvx, \rt=1)$, and, therefore, results in positive latent test error (proposition~\ref{prop:ood_error}).
\\\\
 The \textit{Information Bottleneck} criterion \citep{Tishby2015} introduces a regularization term $\mathcal{R}(q_{\rvz|\rvx})=I_{t=1}(\rvx;\rvz)$ to control the amount of information in the representation, defining a lossy compression scheme \citep{Alemi2016} that can be regulated using the hyper-parameter $\lambda$.
 Note that this criterion does not use any environmental information and blindly discards data-features depending on the regularization strength $\lambda$.
As shown in the example reported in figure~\ref{fig:example_graph}, discarding information without any additional constraint can increase the amount of latent concept shift depending on the structure of the underlying data generating process. This is because discarding information is a necessary but not sufficient condition to reduce the latent concept shift.

\subsection{Independence Criterion}
Whenever the data selection $\rt$ depends only on some observed variable $\rve$ ($\rt=f(\rve;\epsilon)$), the most intuitive approach to reduce the latent test error is to make the representation $\rvz$ independent of the environmental factors $\rve$.
This can be done by minimizing mutual information between $\rve$ and $\rvz$: $\mathcal{R}(q_{\rvz|\rvx}) :=  I_{t=1}(\rve;\rvz)$.
This criterion, known as \textit{independence} or \textit{statistical parity} in the fair classification literature \citep{Dwork2011, Corbett2017}, aims to remove any environmental information from $\rvz$, resulting in a representation that satisfies $p(\rvz|\rt=1) = p(\rvz|\rve,\rt=1)$.

Despite the usefulness of this criterion in the fairness and differential privacy literature, enforcing independence does not necessarily reduce the test error \citep{Zhao2019, Johansson2019}. This is because 
a consistent marginal across different environments does not imply a consistent predictive distribution ($p(\rvy|\rvz,\rve,\rt=1)\neq p(\rvy|\rvz,\rt=1)$), and enforcing independence may even increase the latent concept shift and the test error (as shown in figure~\ref{fig:cmnist_res}).

\subsection{Sufficiency Criterion}
Instead of enforcing a property on the marginal feature distribution, one can consider stable properties of the joint distribution of features $\rvz$ and labels $\rvy$.
The requirement of creating a representation that yields a stable classifier for different values of the environmental factor $\rve$ can be captured by the following regularization: $ \mathcal{R}(q_{\rvz|\rvx}):= I_{t=1}(\rvy;\rve|\rvz)$.
Intuitively minimizing $I_{t=1}(\rvy;\rve|\rvz)$ corresponds to minimizing the distance between the predictive distribution $p(\rvy|\rve,\rvz, \rt=1)$ for each one of the observed environmental conditions.
We can show the following:
\begin{prop}
\label{prop:suff}
Whenever the selection $\rt$ can be expressed as a function of $\rve$ and some independent noise $\epsilon$, the latent concept shift induced by a representation $\rvz$ of $\rvx$ is upper-bounded by $I(\rvy;\rve|\rvz)$:
\begin{align}
    \exists f:\ \rt = f(\rve,\epsilon)
    \implies I(\rvy;\rt|\rvz) \le I(\rvy;\rve|\rvz) .
\end{align}
\end{prop}
In other words, for a given selection $\rt$, it is possible to remove the effect of concept shift by enforcing the \textit{sufficiency} constraint using the variable (or variables) $\rve$ that are responsible for that selection. 

The result from proposition~\ref{prop:suff} is applicable only if the two following conditions are met:
(i) a sufficient representation must exist \citep{Koyama2020}; (ii) enforcing sufficiency on the selected train distribution results in sufficiency for the overall joint distribution ($I_{\rt=1}(\rvy;\rve|\rvz) = 0 \implies I(\rvy;\rve|\rvz) = 0$) \citep{Rosenfeld2020}.
Even if assumptions (i) and (ii) are often acceptable in practice, lack of environment variety at training time or direct dependencies between environmental factors and targets (such as in the y-CMNIST dataset in section~\ref{sec:datasets}) can compromise the effectiveness of the sufficiency criterion. An in depth discussion with a simple example is reported in appendix~\ref{app:pred_info}.

\subsection{Separation Criterion}
The last family of objectives and design principles includes approaches that  aim to capture the stability in the latent feature distribution when the target is observed across different environmental conditions $p(\rvz|\rvy)=p(\rvz|\rvy,\rve)$ \citep{Chouldechova2016, YLi2018}. 
This requirement can be enforced by minimizing the dependency between environmental factors and the representation when the label is observed: $\mathcal{R}(q_{\rvz|\rvx}):= I_{t=1}(\rve;\rvz|\rvy)$.
The resulting \textit{separation} criterion (since $\rvy$ separates $\rvz$ and $\rve$) can be used to identify stable properties of the joint distributions even when the selection $\rt$ depends on both targets $\rvy$ and environmental factors $\rve$:
\begin{prop}
\label{prop:sep}
If the selection $\rt$ can be expressed as a function of $\rve$, $\rvy$ and some independent noise $\epsilon$, the latent concept shift of a representation $\rvz$ of $\rvx$  is upper-bounded by the sum of prior-shift $I(\rvy;\rt)$ and $I(\rve;\rvz|\rvy)$:
\begin{align}
        \exists f:\ \rt = f(\rve,\rvy,\epsilon)
    \implies I(\rvy;\rt|\rvz) \le I(\rvy;\rt)+ I(\rve;\rvz|\rvy).
\end{align}
\end{prop}
When selection $\rt$ and targets $\rvy$ are marginally independent, proposition~\ref{prop:sep} guarantees that the latent concept shift of a representation that enforces separation is zero. Furthermore, whenever the marginal distribution $p(\rvy|\rt=0)$ is known, it is possible to adjust the prediction of the latent classifier on the test distribution\footnote{Further details  on the re-weighting procedure can be found in appendix~\ref{app:prior_shift}}.

Similarly to the other criteria, one needs to assume that enforcing separation on train ($I_{t=1}(\rve;\rvz|\rvy)=0$) suffices to guarantee $I(\rve;\rvz|\rvy)=0$. 
Although a representation that enforces separation always exists, the effectiveness of the separation criteria depends on the data-generating process, since the requirement could be exclusively satisfied by a constant representation.

\section{Experiments}
\label{sec:experiments}

\begin{figure}[!t]
    \centering
    \begin{subfigure}{0.5\textwidth}
    \centering
    \resizebox{.95\textwidth}{!}{
    \input{figures/graphical_models}
    }
    \caption{}
    \label{fig:cmnist_graph}
    \end{subfigure}
    \begin{subfigure}{0.45\textwidth}
    \centering
    \includegraphics[width=\columnwidth,
    trim=0 0.25cm 0 0.2cm, clip
    ]{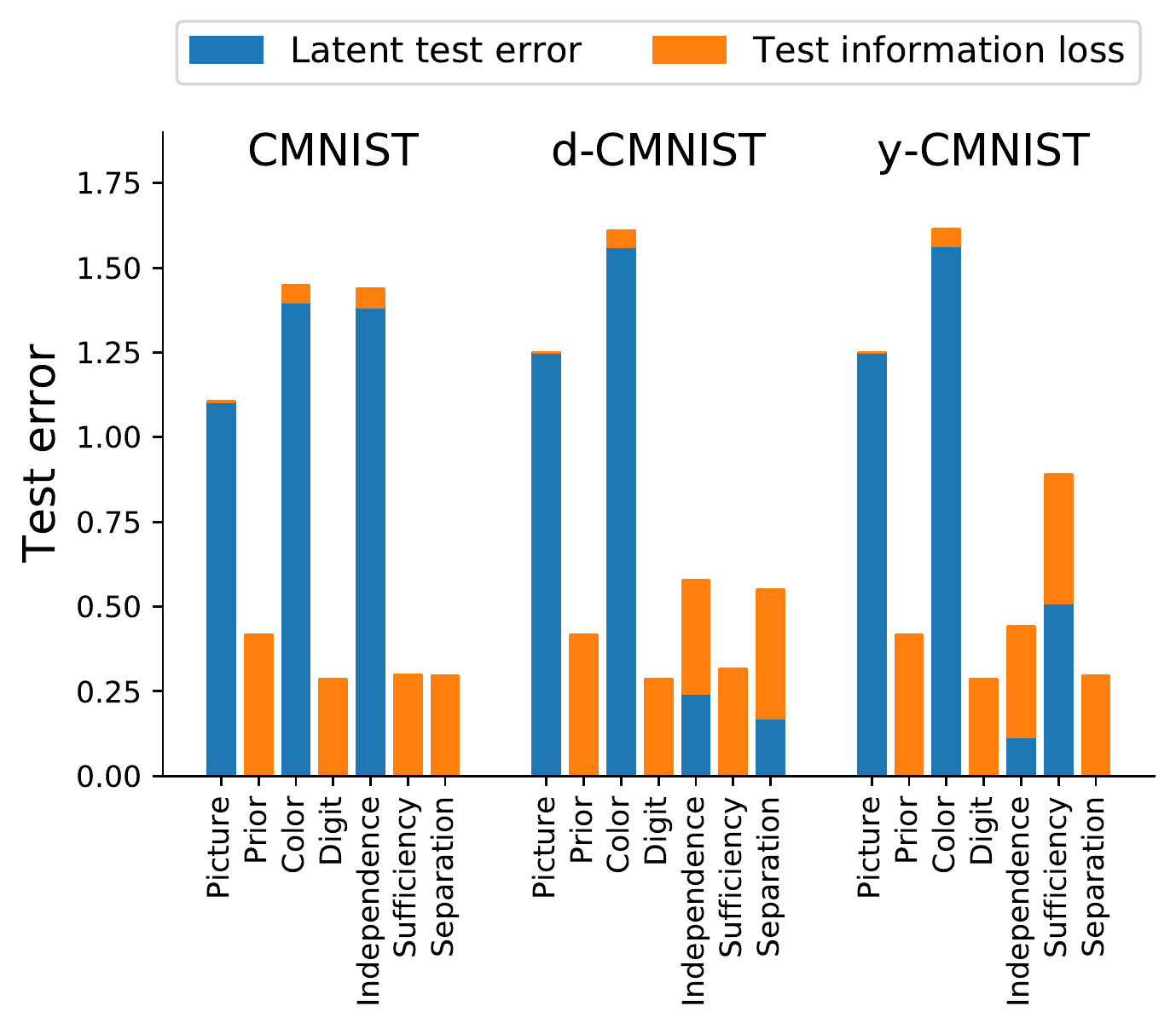}
    \vspace{-1cm}
    \caption{}
    \label{fig:bars}
    \end{subfigure}
    \caption{Graphical models (a) and error components (b) for the CMNIST, d-CMNIST and y-CMNIST data distributions. (a) 
    Dashed lines are used to underline marginal independence between color $\rc$ and environment $\re$, while red arrows denote dependencies added to the original CMNIST distribution. (b) Models trained with strong regularization ($\lambda\approx 10^7$) for the different criteria are compared against the classifiers trained using only color, digit, picture, or prior information. 
    The colors show the proportion of the test error (in nats) due to the predictive information loss ($I_{t=0}(\rvx;\rvy|\rvz)$,
    in orange) and the latent test error ($\KL(p_{\rvy|\rvz}^{\rt=0}||q_{\rvy|\rvz})$,
    in blue) according to the decomposition in equation~\ref{eq:latent_ood}. 
    }
\end{figure}

We evaluate the effectiveness of the criteria presented in section~\ref{sec:criteria} and some of their most popular implementations on multiple versions of the CMNIST dataset \citep{Arjovsky2019} produced by altering the data-generating process (figure~\ref{fig:cmnist_graph}) to underline the shortcomings of the different methods.

The main advantage of using a synthetic dataset such as CMNIST lies in the possibility to directly optimize for the different criteria using differentiable discrete mutual information measurements. This is possible since the joint occurrence of color and digit $\hat\rvx:=[\rc,\rd]$ is a low-dimensional discrete sufficient statistic of the pictures $\rvx$, and, consequently, all the mutual information quantities of interest involving $\rvx\in[0,1]^{28\times28\times2}$ can be computed using $\hat\rvx\in\{0,1\}\times\{0,\hdots,9\}$ instead.
Further details regarding the direct optimization of the criteria are reported in appendix~\ref{app:mi_opt}.

In figure~\ref{fig:cmnist_res}, the theoretical performance for each criterion is compared against the results obtained by training different models on the pictures $\rvx$ using neural network architectures to parametrize the encoder $q(\rz|\rvx)$ and classifier $q(\ry|\rvz)$ for different regularization strength $\lambda$.
For the comparison, we consider diverse popular models designed according to the criteria defined in section~\ref{sec:criteria}:
\begin{itemize}
    \item Variational Information Bottleneck (VIB) \citep{Alemi2016}: a variational tractable approximation of the \textbf{Information Bottleneck} criterion;
    \item Domain Adversarial Neural Network (DANN) \citep{Ganin2015}: an adversarial model based on a min-max game with discriminator $d(\rve|\rvz)$ that is optimized to predict environment information from the representation $\rvz$ to enforce the \textbf{Independence} criterion;
   \item Invariant Risk Minimization (IRM) \citep{Arjovsky2019}: A model designed following the \textbf{Sufficiency} criterion that aims to create a representation from which the same latent classifier is simultaneously optimal on all environments.
    \item Conditional Domain Adversarial Neural Network (CDANN) \citep{YLi2018}: an adversarial approximation of the \textbf{Separation} criterion in which, analogously to DANN, a discriminator tries to predict the environment when the representation $\rvz$ and the true label $\rvy$ are given.
     \item Variance-based Risk Extrapolation (VREx) \citep{Krueger2020}: a model designed following the \textbf{Sufficiency} and \textbf{Independence} criteria, based on the minimization of the training error variance across the different environments;
\end{itemize}
For a better comparison, all the models use the same encoder and classifier neural network architectures. Each model has been trained by slowly increasing the regularization strength after an initial pre-training with small $\lambda$. Further details regarding the neural network architectures, objectives, optimization and specific hyper-parameters can be found in appendix~$\ref{app:nn_train}$.\footnote{The implementation of the models in this work is available at \url{https://github.com/mfederici/dsit}}

\begin{figure*}[!t]
    \centering
    \includegraphics[width=0.95\textwidth]{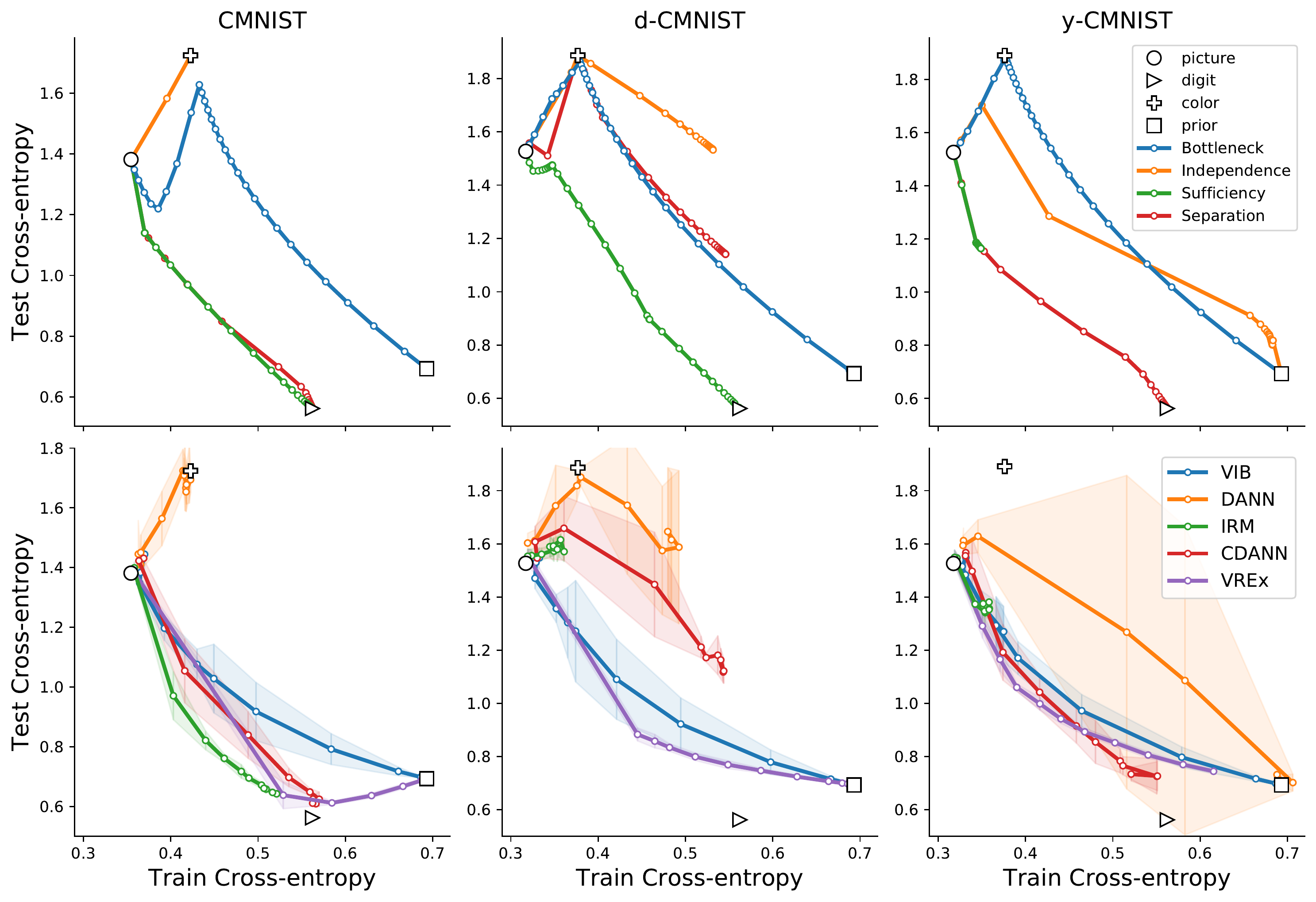}
    \caption{Comparison of training cross-entropy (x-axis) and test cross-entropy (y-axis) on the CMNIST, d-CMNIST and y-CMNIST datasets for representations trained using different criteria directly (top row) and corresponding approximations/model (bottom row). Each trajectory describes the trade-off between the two errors for a wide range of regularization strength values $\lambda$.
    The white symbols in each plot represent the error of models that consider only color ($p(\ry|\rc,\rt=1)$), digit ($p(\ry|\rd,\rt=1)$), picture ($p(\ry|\rvx,\rt=1)$) or prior  ($p(\ry|\rt=1)$) information.
    The shaded area represents the standard deviation observed across three runs with different seeds.
    }
    \label{fig:cmnist_res}
\end{figure*}

\subsection{Evaluation metric} 
The measure of test error, information loss and latent test error reported in figure~\ref{fig:bars} can be computed only for discrete variables.
For neural network models the training and test error defined in section~\ref{sec:framework} can be estimated up to a constant entropy by considering the expected negative log-likelihood (empirical cross-entropy):
\begin{align}
    \KL(p_{\rvy|\rvx}^{\rt=1}||q_{\rvy|\rvx})
    \approx -\frac{1}{N}\sum_{i=1}^N \log q(\rvy=\vy_i|\rvx=\vx_i) - \underbrace{H_{\rt=1}(\rvy|\rvx)}_{\text{constant}}
\end{align}
The samples $\vx_i,\vy_i$ are obtained from $p(\rvx,\rvy|\rt=1)$ and $p(\rvx,\rvy|\rt=0)$ for train and test cross-entropy respectively.
By computing the training and test cross-entropy values for different regularization strength $\lambda$, each model defines a trajectory from a maximum likelihood solution ($\lambda=0$) to the results obtained when the corresponding independence constraint is enforced (large $\lambda$), as shown in figure~\ref{fig:cmnist_res}.
Contrarily to accuracy measurements, the use of cross-entropy allows us to detect under-confident or over-confident predictive distributions. 

\subsection{Datasets}
\label{sec:datasets}

The two variants of the CMNIST dataset considered in this work have been designed by minimally changing the original distribution to underline the strengths and weaknesses of the different criteria. 
Across different experiments, each MNIST picture $\rvx$ of a digit $\rd$ is associated with a color $\rc$ that depends on a binary target $\ry$ and an environment $\re$. The d-CMNIST and y-CMNIST datasets are created by adding a dependency from the environment to digit $\rd$ (d-CMNIST) and label $\ry$ (y-CMNIST), which results in a correlation between targets and environment ($I(\re;\ry)>0$). Further details regarding the conditional distributions used to produce the different datasets can be found in appendix~\ref{app:cmnist_probs}.

The strong correlation between color and label across the different CMNIST versions can be seen as an artifact introduced by the selection $\rt=f(\re)$, and models that consider only digit information (white triangles in figure~\ref{fig:cmnist_res}) outperform the ones that capture color information (white crosses) or both (white circle) in terms of test cross-entropy (y-axis). 

\paragraph{CMNIST} 
The CMNIST dataset was originally designed to underline the weaknesses of maximum-likelihood and Empirical Risk Minimization strategies
\citep{Arjovsky2019}.
The results displayed in the first column of figure~\ref{fig:cmnist_res} confirm that both the sufficiency and separation criteria manage to effectively reduce the test cross entropy error for sufficiently strong regularization $\lambda$, 
 minimizing the latent test error while retaining more predictive information than a constant representation (first column in figure~\ref{fig:bars}).
On the CMNIST dataset,  most of the models in analysis closely follow the trajectory estimated by optimizing the corresponding criterion directly. The trajectory defined by the VIB model differs from the one described by the Information Bottleneck criterion since, in absence of additional constraints, the nature of the information that is discarded (either color, digit or style) depends on the inductive bias introduced by the specific architecture.

\paragraph{d-CMNIST} The d-CMNIST dataset adds a dependency between environment $\re$ and digit $\rd$, increasing the frequency of specific digits for some environments. 
Digit information makes environment and label conditionally independent ($\ry$ and $\re$ are d-separated in figure~\ref{fig:cmnist_graph}). As claimed in \citet{Arjovsky2019}, models based on the sufficiency criterion manage to reduce the latent test error (second column of figure~\ref{fig:bars}), while the separation criterion fails because of the direct dependency between environment and digits. The independence criterion does not improve the test performance since both color and digit correlate with the environment. Both DANN and CDANN architectures results in trajectories that are similar to the ones obtained optimizing for the corresponding criteria. 
Note that enforcing separation or independence does not improve upon the result that can be obtained by blindly discarding information using VIB.
Despite the effectiveness of the sufficiency criterion, the model trained with the IRM objective lies far from the optimal solution. We believe that this is due to the relaxations and approximation introduced by the optimized objective as discussed in appendix~\ref{app:empirical_problems}.

\paragraph{y-CMNIST} Adding a dependency between environment $\re$ and label $\ry$ simulates the scenario in which some labels are more frequent in some environments. The arrow between $\rd$ and $\ry$ flips when compared to the d-CMNIST dataset to represent stable $p(\rd|\ry)$ across different environments. The corresponding y-CMNIST dataset includes a path from $\re$ to $\ry$ that can not be blocked since no representation $\rvz$ can achieve sufficiency ($I(\re;\ry|\rvz)=0$).
Despite the optimality of the separation criterion (red line, third column of figure~\ref{fig:cmnist_res}), the model trained with the CDANN objective struggles to minimize the test error due to instabilities of the adversarial training procedure, as discussed in appendix~\ref{app:empirical_problems}.
Once again, the trajectory described by the VREx objective is more favorable when compared to VIB, while the model trained using the IRM objective is far from optimality.
Figure~\ref{fig:bars} confirms that, on the y-CMNIST dataset, the separation criterion is the only one that minimizes the latent test error without discarding the entirety of the picture information.
This underlines that the effectiveness of sufficiency and separation criteria strongly depends on the structure of the underlying graphical model.

\section{Related work}
The problem of distribution (or dataset) shift \citep{Quionero2009, Koh2020} has been explored in different areas in the machine learning literature ranging  domain adaptation and generalization \citep{Wang2018} to sub-population shift and fair classification \citep{Mehrabi2019, Santurkar2020}.
Although goals, availability of test features and environments at training time, and data selection criteria \citep{Koh2020} may differ, most recent approaches focus on extracting features with some desired properties through three different families of objectives.

The independence criterion has been used to find the transformation which minimizes the distance between the distribution of the encoded features across different environments by using adversarial training \citep{Ganin2015,Xie2017,Li2018}, features adjustment \citep{Lum2016}, kernel methods \citep{Muandet2013}, or variational approaches \citep{Louizos2015, Moyer2018, Ilse2019}. Despite the success of the independence criterion, \citet{Zhao2019, Johansson2019} have shown that enforcing stability of the marginal feature distribution is not sufficient to guarantee generalization.

\citet{YLi2018} extends the independence criterion using adversarial training to create representations that are independent of the environment when the label is observed. The corresponding separation criterion has been explored in the context of fair classification as \textit{equalized odds} \citep{Hardt2016} and some tractable relaxations \citep{Darlington1971, Woodworth2017,Chouldechova2016}.

The idea of considering features that lead to a consistent predictive distribution has been explored in the causality literature \citep{Peters2015, Magliacane2017} for linear models as a feature selection criteria. 
Other work approached the problem in the non-linear case by considering gradients of the classifier \citep{Arjovsky2019, Koyama2020, Parascandolo2020} or penalizing the variance of error \citep{Krueger2020, Xie2020} across different environments.
Other relaxations of the sufficiency criterion such as as \textit{predictive parity} \citep{Chouldechova2016} have been considered in the fairness literature. 

Despite the promising directions of research, \citet{Gulrajani2020} have shown that the performance of the most popular approaches in the literature strongly depends on the hyper-parameter tuning strategy, underlying the problem of the lack of (i) standardized benchmark procedures and (ii) clarity regarding guarantees and hidden assumptions for the different algorithms. Although the problem of distribution shift has been widely studied in the classic domain generalization literature \citep{Mansour2009, Scholkopf2012, Gong2016}, this work aims to characterize the problem from a representation learning and information theoretical perspective, providing different insights and identifying potential issues  and benefits of several popular models.

\section{Discussion and Conclusion}
In this work, we characterize an information-theoretic framework to analyze the distribution shift problem by relating it to the train and out of distribution test error. We identify two sources of errors for models based on latent representations, and we show that different criteria explored in literature can be seen as different strategies to minimize concept shift in the latent space using extra information about the data selection procedure.
We demonstrate both theoretically and empirically that their effectiveness depends on the structure of the underlying graphical model with respect to the observed variables, training data, and data selection criteria other than the chosen approximation and relaxations for optimization. 

Although test information loss and latent test error are challenging to estimate for real-world datasets, we argue that the presented analysis can be useful to better understand and mitigate the effect of selection bias and distribution shift in machine learning models. An accurate estimation of the error decomposition reported in this work could be further used to guide processes of causal discovery.


\begin{ack}
We thank Sindy L\"{o}we and David Ruhe for their insightful comments and feedback. This work was supported by the Microsoft Research PhD Scholarship Programme.
\end{ack}

\bibliography{bibliography}
\bibliographystyle{icml2021}

\newpage
\appendix

\section{Notation}
\label{app:notation}
For a given joint distribution $p(\rvx,\rvy,\rve)$ on $\rvx$, $\rvy$ and $\rve$, a binary selection $\rt=f(\rvx,\rvy,\rve,\epsilon)$, and a conditional distribution $q(\rvy|\rvx)$ we use the following notation

\subsection{Conditional Kullback-Leibler divergence}
\begin{align}
    \KL(p(\rvy|\rvx)||q(\rvy|\rvx)):=\E_{\vx,\vy\sim p(\rvx,\rvy)}\left[ \log\frac{p(\rvy=\vy|\rvx=\vx)}{q(\rvy=\vy|\rvx=\vx)}\right].
\end{align}
Note that the expectation is always considered with respect to the joint distribution for the first argument of the KL-divergence.
The notation $\KL(p_{\rvy|\rvx}||q_{\rvy|\rvx})$ will be used to abbreviate the same quantity.
\subsection{Jensen-Shannon divergence}
\begin{align}
    \JSD(p(\rvx)||q(\rvx))&:=\frac{1}{2} \KL(p(\rvx)||m(\rvx))+ \frac{1}{2}\KL(q(\rvx)||m(\rvx)),
    \label{def:jsd}
\end{align}
with $m(\rvx) = 1/2\ p(\rvx) + 1/2\ q(\rvx)$.

\subsection{Mutual Information}
\begin{align}
    I(\rvx;\rvy) &:= \KL(p(\rvx,\rvy)||p(\rvx)p(\rvy))\\
    &=\KL(p(\rvy|\rvx)||p(\rvy))\\
    &=\KL(p(\rvx|\rvy)||p(\rvx)).
\end{align}
The subscript $\rt=1$ is used to indicate that both joint and marginal distribution are conditioned on $\rt=1$:
\begin{align}
    I_{\rt=1}(\rvx;\rvy) &:= \KL(p(\rvx,\rvy|\rt=1)||p(\rx|\rt=1)p(\rvy|\rt=1))
\end{align}
Conditional mutual information is defined as:
\begin{align}
    I(\rvx;\rvy|\rve) &:= \KL(p(\rvx,\rvy|\rve)||p(\rvx|\rve)p(\rvy|\rve))\\
    &:= \KL(p(\rvy|\rvx,\rve)||p(\rvy|\rve))\\
    &:= \KL(p(\rvx|\rvy,\rve)||p(\rvx|\rve)).
\end{align}
Note that all the mutual information terms in this work are implicitly expressed in terms of the distribution $p$.

\section{Proofs and Additional details}
\label{app:proofs}
The proofs for the statement in the main text are reported in this section.


\subsection{Proof for Proposition~\ref{prop:ood_error}}
\begin{proof}
    Consider $\KL(p(\rvy|\rvx)||q(\rvy|\rvx))$. By multiplying and dividing by $p(\rvy|\rvx,\rt)$, we obtain:
    \begin{align}
        \KL(p(\rvy|\rvx)||q(\rvy|\rvx))&=\E_{\vx,\vy\sim p(\rvx,\rvy)}\left[ \log\frac{p(\rvy=\vy|\rvx=\vx)}{q(\rvy=\vy|\rvx=\vx)}\right]\nonumber\\
        &=\E_{\vx,\vy,t\sim p(\rvx,\rvy,\rt)}\left[ \log\frac{p(\rvy=\vy|\rvx=\vx,\rt=t)}{q(\rvy=\vy|\rvx=\vx)}\frac{p(\rvy=\vy|\rvx=\vx)}{p(\rvy=\vy|\rvx=\vx,\rt=t)}\right]\nonumber\\
        &=\KL(p(\rvy|\rvx,\rt)||q(\rvy|\rvx)) - I(\rvy;\rt|\rvx).\label{eq:prop_1_1}
    \end{align}
    Splitting the expectation on $\rt$, we express the first term as:
    \begin{align}
    \KL(p(\rvy|\rvx,\rt)||q(\rvy|\rvx))&=p(\rt=1) \E_{\vx,\vy\sim p(\rvx,\rvy|\rt=1)}\left[ \log\frac{p(\rvy=\vy|\rvx=\vx,\rt=1)}{q(\rvy=\vy|\rvx=\vx)}\right]\nonumber\\
        &\ \ \ + p(\rt=0)\ \E_{\vx,\vy\sim p(\rvx,\rvy|\rt=0)}\left[ \log\frac{p(\rvy=\vy|\rvx=\vx,\rt=0)}{q(\rvy=\vy|\rvx=\vx)}\right]\nonumber\\
        &=p(\rt=1)\KL(p(\rvy|\rvx,\rt=1)||q(\rvy|\rvx)) + p(\rt=0)\KL(p(\rvy|\rvx,\rt=0)||q(\rvy|\rvx))\nonumber\\
        &=p(\rt=1)\KL(p_{\rvy|\rvx}^{\rt=1}||q_{\rvy|\rvx})+p(\rt=0)\KL(p_{\rvy|\rvx}^{\rt=0}||q_{\rvy|\rvx}).\label{eq:prop_1_2}
    \end{align}
    Since the KL-divergence between two distributions is always positive, using the result from equations~\ref{eq:prop_1_1} and ~\ref{eq:prop_1_2}, we have:
    \begin{align}
        I(\rvy;\rt|\rvx) &\le \KL(p(\rvy|\rvx,\rt)||q(\rvy|\rvx))\nonumber\\
        &=p(\rt=1)\KL(p_{\rvy|\rvx}^{\rt=1}||q_{\rvy|\rvx})+p(\rt=0)\KL(p_{\rvy|\rvx}^{\rt=0}||q_{\rvy|\rvx})\nonumber\\
        &\le (1-\alpha)\left(\KL(p_{\rvy|\rvx}^{\rt=1}||q_{\rvy|\rvx})+\KL(p_{\rvy|\rvx}^{\rt=0}||q_{\rvy|\rvx})\right),
        \label{eq:proof_prop1_last}
    \end{align}
    with $\alpha:=\min\{p(\rt=0),p(\rt=1)\}$.
    
    Whenever $\alpha=1$, the inequality in equation~\ref{eq:proof_prop1_last} holds since $I(\rvy;\rt|\rvx)=0$ and KL-divergence is always positive.
    For $\alpha<1$ we can divide both sides by $1-\alpha$, obtaining:
    \begin{align*}
        \KL(p_{\rvy|\rvx}^{\rt=1}||q_{\rvy|\rvx})+\KL(p_{\rvy|\rvx}^{\rt=0}||q_{\rvy|\rvx})\ge \frac{1}{1-\alpha} I(\rvy;\rt|\rvx)
    \end{align*}
\end{proof}

\subsection{Proof for Proposition~\ref{prop:decomposition}}
\begin{proof}
    Using Jensen's inequality we have:
\begin{align}
    \KL(p_{\rvy|\rvx}^{\rt=1}||q_{\rvy|\rvx}) &= \E_{\vx,\vy\sim(\rvx,\rvy|\rt=1)}\left[\log\frac{p(\rvy=\vy|\rvx=\vx,\rt=1}{\E_{\vz\sim q(\rvz|\rvx=\vx)}[q(\rvy=\vy|\rvz=\vz)]}\right]\nonumber\\ 
    &\le^* \E_{\vx,\vy\sim(\rvx,\rvy|\rt=1)}\E_{\rz\sim q(\rvz|\rvx=\vx)}\left[\log\frac{p(\rvy=\vy|\rvx=\vx,\rt=1)}{q(\rvy=\vy|\rvz=\vz)}\right]\nonumber\\
    &=\KL(p_{\rvy|\rvx}^{\rt=1}||q_{\rvy|\rvz}),
    \label{eq:latent_bound_0}
\end{align}
in which $^*$ holds with equality when $q(\rvz|\rvx)$ is a delta distribution (deterministic encoder).

Secondly, by multiplying and dividing by $p(\rvy=\vy|\rvz=\vz, \rt=1)$:
\begin{align}
    \KL(p_{\rvy|\rvx}^{\rt=1}||q_{\rvy|\rvz})
    &= \E_{\vx,\vy\sim(\rvx,\rvy|\rt=1)}\E_{\rz\sim q(\rvz|\rvx=\vx)}\left[\log\frac{p(\rvy=\vy|\rvx=\vx,\rt=1)}{p(\rvy=\vy|\rvz=\vz, \rt=1)}\frac{p(\rvy=\vy|\rvz=\vz, \rt=1)}{q(\rvy=\vy|\rvz=\vz)}\right]\nonumber\\
    &= I(\rvy;\rvx|\rvz) + \KL(p_{\rvy|\rvz}^{\rt=1}||q_{\rvy|\rvz}).\label{eq:latent_bound_1}
\end{align}

The bound for $\KL(p_{\rvy|\rvx}^{\rt=0}||q(\rvy|\rvx))$ is obtained analogously.
\end{proof}

\subsection{Proof for Proposition~\ref{prop:uebound}}
\begin{proof}
    We first define the average latent predictive distribution $\overline p(\rvy|\rvz):=\frac{1}{2}p(\rvy|\rvz,\rt=0)+\frac{1}{2}p(\rvy|\rvz,\rt=1)$, then
    \begin{align}
        I(\rvy;\rt|\rvz) &= \KL(p(\rvy|\rvz,\rt)||p(\rvy|\rvz))\nonumber\\
        &= p(\rt=0) \KL(p_{\rvy|\rvz}^{\rt=0}||p_{\rvy|\rvz})+ p(\rt=1) \KL(p_{\rvy|\rvz}^{\rt=1}||p_{\rvy|\rvz})\nonumber\\
    & \ge 2\alpha\left(\frac{1}{2}\KL(p_{\rvy|\rvz}^{\rt=0}||p_{\rvy|\rvz})+\frac{1}{2} \KL(p_{\rvy|\rvz}^{\rt=1}||p_{\rvy|\rvz})\right)\nonumber\\
        &= 2\alpha\  \E_{t\sim Ber(1/2)}\E_{\vz,\vy\sim p(\rvz,\rvy|\rt=t)}\left[\log\frac{p(\rvy=\vy|\rvz=\vz,\rt=t)}{p(\rvy=\vy|\rvz=\vz)}\right]\nonumber\\
        &= 2\alpha\  \E_{t\sim Ber(1/2)}\E_{\vz,\vy\sim p(\rvz,\rvy|\rt=t)}\left[\log\frac{p(\rvy=\vy|\rvz=\vz,\rt=t)}{\overline p(\rvy=\vy|\rvz=\vz)}\frac{\overline p(\rvy=\vy|\rvz=\vz)}{p(\rvy=\vy|\rvz=\vz)}\right]\nonumber\\
        &= 2\alpha\ \JSD(p^{\rt=0}_{\rvy|\rvz}||p^{\rt=1}_{\rvy|\rvz}) + \KL(\overline{p}(\rvy|\rvz)||p(\rvy|\rvz))\nonumber\\
        &\ge 2\alpha\ \JSD(p^{\rt=0}_{\rvy|\rvz}||p^{\rt=1}_{\rvy|\rvz}),
        \label{eq:theq_1}
    \end{align}
    where $Ber(1/2)$ refers to a Bernoulli distribution with parameter $p=1/2$.
    
    Secondly, since the square root of the Jensen-Shannon divergence is a metric \citep{Endres2003}, using triangle inequality we have:
    \begin{align}
        \sqrt{\JSD(p^{\rt=1}_{\rvy|\rvz=\vz}||q_{\rvy|\rvz=\vz})}+\sqrt{\JSD(p^{\rt=0}_{\rvy|\rvz=\vz}||p^{\rt=1}_{\rvy|\rvz=\vz})} \ge
        \sqrt{\JSD(p^{\rt=0}_{\rvy|\rvz=\vz}||q_{\rvy|\rvz=\vz})}.
        \label{eq:theq_2}
    \end{align}
    Note that the conditional distributions $p^{\rt=0}_{\rvy|\rvz=\vz}$ and $p^{\rt=1}_{\rvy|\rvz=\vz}$ are defined only for latent vectors $\vz$ that have positive probability according to $p(\rvz|\rt=0)$ and $p(\rvz|\rt=1)$ respectively. Therefore inequality is defined only for $\vz$ that have simultaneously strictly positive probability on both train and test distributions.
    
    Lastly, The Jensen-Shannon divergence can be upper-bounded by a function of the corresponding Kullback-Leibler divergence: 
    \begin{align}
        \JSD(p_{\rvy|\rvz}^{\rt=1}||q_{\rvy|\rvz}) = \frac{1}{2} \KL(p_{\rvy|\rvz}^{\rt=1}||q_{\rvy|\rvz})-\KL(m_{\rvy|\rvz}||q_{\rvy|\rvz})\le \frac{1}{2} \KL(p_{\rvy|\rvz}^{\rt=1}||q_{\rvy|\rvz}),
        \label{eq:theq_3}
    \end{align}
    with $m_{\rvy|\rvz} := 1/2 p(\rvy|\rvz,\rt=1)+ 1/2 q(\rvy|\rvz)$.

    Using the results from \ref{eq:theq_1}, \ref{eq:theq_2} and \ref{eq:theq_3} we have
    \begin{align}
        \JSD(p^{\rt=0}_{\rvy|\rvz=\vz}||q_{\rvy|\rvz=\vz}) &\stackrel{(\ref{eq:theq_2})}{\le}\left(\sqrt{\JSD(p^{\rt=1}_{\rvy|\rvz=\vz}||q_{\rvy|\rvz=\vz})}+\sqrt{\JSD(p^{\rt=0}_{\rvy|\rvz=\vz}||p^{\rt=1}_{\rvy|\rvz=\vz})}\right)^2\nonumber\\
        &\stackrel{(\ref{eq:theq_1})}{\le}\left(\sqrt{\frac{1}{2\alpha}I(\rvy;\rt|\rvz=\vz)}+\sqrt{\JSD(p^{\rt=0}_{\rvy|\rvz=\vz}||p^{\rt=1}_{\rvy|\rvz=\vz})}\right)^2\nonumber\\
        &\stackrel{(\ref{eq:theq_3})}{\le}\left(\sqrt{\frac{1}{2\alpha}I(\rvy;\rt|\rvz=\vz)}+\sqrt{\frac{1}{2}\KL(p_{\rvy|\rvz=\vz}^{\rt=1}||q_{\rvy|\rvz=\vz})}\right)^2
        \label{eq:boundproof}
    \end{align}
\end{proof}
\subsection{Upper-bounding the test error}
We show two inequalities, that relate the Jensen-Shannon divergence in equation \ref{eq:boundproof} to the latent test error (defined as a Kullback-Leibler divergence).

\subsubsection{Ratio of $f$-divergences}
Since both Jensen-Shannon and Kullback Leibler divergence are instances of $f$-divergences, we can use the result from theorem 1 in \citet{Sason2016} to write the following inequality:

\begin{align}
    \JSD(p^{\rt=0}_{\rvy|\rvz=\vz}||q_{\rvy|\rvz=\vz}) \ge K(p^{\rt=0}_{\rvy|\rvz=\vz},q_{\rvy|\rvz=\vz}) \KL(p^{\rt=0}_{\rvy|\rvz=\vz}||q_{\rvy|\rvz=\vz}),
\end{align}
with 
\begin{align*}
    K(p^{\rt=0}_{\rvy|\rvz=\vz},q_{\rvy|\rvz=\vz}) = \sup_{\vy} k\left(\frac{p(\rvy=\vy|\rvz=\vz,\rt=0)}{q(\rvy=\vy|\rvz=\vz)}\right),
\end{align*}
in which
\begin{align*}
    k(r) = \begin{cases}\frac{f(r)}{g(r)} \ \ \ \text{ for } r\in(0,1)\cup(1,+\infty)\\
    1 \ \ \ \ \ \ \ \ \text{ for } r=1
    \end{cases},
\end{align*}
and
\begin{align*}
    f(r) &= r\log r\\
    g(r) &= \frac{1}{2}\left[r\log r - (r+1)\log\frac{r+1}{2}\right]
\end{align*}
\subsubsection{Pinsker-type bound}
Fisrt, since the Jensen-Shannon divergence is a $f$-divergence, we can use the result from theorem~3 in \citet{Gilardoni2010} to express a lower-bound as a function of the total variation $D_{TV}$:
\begin{align}
    \sqrt{\JSD(p^{\rt=0}_{\rvy|\rvz=\vz}||q_{\rvy|\rvz=\vz})}&\ge \sqrt{\frac{f''(1)}{2}}\ D_{TV}(p^{\rt=0}_{\rvy|\rvz=\vz}||q_{\rvy|\rvz=\vz})\nonumber\\
    &=\sqrt{2}\ D_{TV}(p^{\rt=0}_{\rvy|\rvz=\vz}||q_{\rvy|\rvz=\vz}),
\end{align}
with $f''(1)$ as the second derivative of $f(t) = \frac{1}{2}\left[(t+1)\log\frac{2}{t+1} + t\log t\right]$ evaluated in $t=1$.

Secondly, since the Kullback-Leibler divergence is also a $f$-divergence, we use the result of theorem~1 from \citet{Binette2019} to upper-bound the latent test error as a function of the total variation:
\begin{align}
    \KL(p^{\rt=0}_{\rvy|\rvz=\vz}||q_{\rvy|\rvz=\vz})&\le \left(\frac{f(m(\rvz))}{1-m(\rvz)}+\frac{f(M(\rvz))}{M(\rvz)-1}\right) D_{TV}(p^{\rt=0}_{\rvy|\rvz=\vz}||q_{\rvy|\rvz=\vz})\nonumber\\
    &= \left(\frac{m(\vz)\log m(\vz)}{1-m(\vz)}+\frac{M(\vz)\log M(\vz)}{M(\vz)-1}\right) D_{TV}(p^{\rt=0}_{\rvy|\rvz=\vz}||q_{\rvy|\rvz=\vz}),
\end{align}
with $m(\vz):=\min_\vy\frac{p(\rvy=\vy|\rvz=\vz, t=0)}{q(\rvy=\vy|\rvz=\vz)}$, $M(\vz):=\max_\vy\frac{p(\rvy=\vy|\rvz=\vz, t=0)}{q(\rvy=\vy|\rvz=\vz)}$, and $f(t) = t\log t$.

Combining the two bounds we obtain:
\begin{align}
    \KL(p^{\rt=0}_{\rvy|\rvz=\vz}||q_{\rvy|\rvz=\vz})\le \frac{1}{\sqrt{2}}\left(\frac{m(\rvz)\log m(\vz)}{1-m(\vz)}+\frac{M(\vz)\log M(\vz)}{M(\vz)-1}\right) \sqrt{\JSD(p^{\rt=0}_{\rvy|\rvz=\vz}||q_{\rvy|\rvz=\vz})}.
\end{align}
Note that the right-hand side of this last inequality is finite whenever the support of the model $q(\rvy|\rvz=\vz)$ contains the support of $p(\rvy|\rvz=\vz, \rt=0)$ (and therefore $M(\vz)$ is finite). When this does not happen, both left-hand and right hand side of the inequality are unbounded.

\subsection{Bound in expression~\ref{eq:bounds_info}}
\begin{proof}
Consider the amount of predictive information lost by encoding $\rvx$ into $\rvz$:
\begin{align}
     I_{\rt=1}(\rvx;\rvy|\rvz)&= I_{\rt=1}(\rvz\rvx;\rvy)-I_{\rt=1}(\rvz;\rvy)\nonumber\\
     &=I_{\rt=1}(\rvx;\rvy)+I_{\rt=1}(\rvz;\rvy|\rvx)-I_{\rt=1}(\rvz;\rvy)\nonumber\\
     &=I_{\rt=1}(\rvx;\rvy)-I_{\rt=1}(\rvz;\rvy),
\end{align}
in which $I_{\rt=1}(\rvz;\rvy|\rvx)=0$ since $\rvz$ depends only on $\rvx$.

Since $I(\rvx;\rvy)$ is constant in $q(\rvz|\rvx)$:
\begin{align}
    \min_{q(\rvz|\rvx)}I_{\rt=1}(\rvx;\rvy|\rvz) = I(\rvx;\rvy) - \max_{q(\rvz|\rvx)} I_{\rt=1}(\rvy;\rvz).
\end{align}
The upper-bound is derived considering $q(\rvy|\rvz)$ as a variational distribution.
For any $q(\rvy|\rvz)$:
\begin{align}
    I_{\rt=1}(\rvx;\rvy|\rvz) &= \KL(p_{\rvy|\rvx}^{\rt=1}||p_{\rvy|\rvz}^{\rt=1})\nonumber\\
    &= \KL(p_{\rvy|\rvx}^{\rt=1}||q_{\rvy|\rvz}) - \KL(p_{\rvy|\rvz}^{\rt=1}||q_{\rvy|\rvz})\nonumber\\
    &\le \KL(p_{\rvy|\rvx}^{\rt=1}||q_{\rvy|\rvz}),
\end{align}
and in particular
\begin{align}
    I_{\rt=1}(\rvx;\rvy|\rvz) &\le \min_{q(\rvy|\rvz)} \KL(p_{\rvy|\rvx}^{\rt=1}||q_{\rvy|\rvz}),
\end{align}
in which equality holds for $q(\rvy|\rvz)=p(\rvy|\rvz,\rt=1)$.
\end{proof}

\subsection{Proof for Proposition~\ref{prop:suff}}
\begin{proof}
    Consider the latent concept shift $I(\rvy;\rt|\rvz)$. Using the chain rule of mutual information, we obtain:
    \begin{align}
        I(\rvy;\rt|\rvz) &= I(\rvy;\rve\rt|\rvz)-I(\rvy;\rve|\rvz\rt)\nonumber\\
        &\le I(\rvy;\rve\rt|\rvz)\nonumber\\
        &= I(\rvy;\rve|\rvz) + I(\rvy;\rt|\rvz\rve).
    \end{align}
    Whenever $\rt$ can be expressed as a function of $\rve$ and some independent noise $\epsilon$, we have:
    \begin{align}
        &\rt=f(\rve,\epsilon)\nonumber\\
        &\implies p(\rt|\rve) = p(\rt|\rve,\rvz) =  p(\rt|\rve,\rvz,\rvy)\nonumber\\
        &\implies \KL(p(\rt|\rve,\rvz)||p(\rt|\rve,\rvz,\rvy))=0\nonumber\\
        &\iff I(\rvy;\rt|\rvz\rve)=0.
    \end{align}
    Therefore, we conclude:
    \begin{align}
        I(\rvy;\rt|\rvz)\le I(\rvy;\re|\rvz)
    \end{align}
\end{proof}

\subsection{Proof for Proposition~\ref{prop:sep}}
\begin{proof}
    Consider the latent concept shift $I(\rvy;\rt|\rvz)$. Using the chain rule of mutual information, we obtain:
    \begin{align}
        I(\rvy;\rt|\rvz) &= I(\rvy\rvz;\rt) - I(\rt;\rvz)\nonumber\\
        &\le I(\rvy\rvz;\rt)\nonumber\\
        &= I(\rvy;\rt) + I(\rt;\rvz|\rvy)
        \label{eq:prop_sep_1}
    \end{align}
    Then, we express $I(\rt;\rvz|\rvy)$ as a function of $I(\rve;\rvz|\rvy)$:
    \begin{align}
        I(\rt;\rvz|\rvy) &= I(\rve\rt;\rvz|\rvy)-I(\rve;\rvz|\rvy\rt) \nonumber\\
        &\le I(\rve\rt;\rvz|\rvy)\nonumber\\
        &= I(\rve;\rvz|\rvy) + I(\rt;\rvz|\rvy\rve).
        \label{eq:prop_sep_2}
    \end{align}
    
    Whenever $\rt$ can be expressed as a function of $\rve$, $\rvy$ and some independent noise $\epsilon$, we have:
    \begin{align}
        &\exists f:\ \rt = f(\rve,\rvy,\epsilon)\nonumber\\
        &\implies p(\rt|\rve,\rvy) = p(\rt|\rve,\rvy,\rvz)\nonumber\\
        &\implies \KL(p(\rt|\rve,\rvy,\rvz)||p(\rt|\rve,\rvy))=0\nonumber\\
        &\iff I(\rt;\rvz|\rvy\rve)=0.
        \label{eq:prop_sep_3}
    \end{align}
    
    Using the results from equations~\ref{eq:prop_sep_1}, \ref{eq:prop_sep_2} and \ref{eq:prop_sep_3}:
    \begin{align}
        I(\rvy;\rt|\rvz) &\le  I(\rvy;\rt) + I(\rt;\rvz|\rvy)\nonumber\\
        &\le I(\rvy;\rt) + I(\rve;\rvz|\rvy).
    \end{align}
\end{proof}

\subsection{Correcting for prior shift}
\label{app:prior_shift}
When the separation constraint $I(\rve;\rvz|\rvy)=0$ is enforced the \textit{reverse concept shift} $I(\rt;\rvz|\rvy)$, which represents how much the distribution $p(\rvz|\rvy)$ changes as a result of the selection, is zero (equations~\ref{eq:prop_sep_2} and \ref{eq:prop_sep_3}).
Considering the stability of $p(\rvz|\rvy)$, we can express the test predictive distribution $p(\rvy|\rvz,\rt=0)$ as a function of the train one using Bayes rule:
\begin{align}
    p(\rvy|\rvz,\rt=0) &= \frac{p(\rvz|\rvy,\rt=0)p(\rvy|\rt=0)}{p(\rvz|\rt=0)}\nonumber\\ 
    &=^* \frac{p(\rvz|\rvy,\rt=1)p(\rvy|\rt=0)}{p(\rvz|\rt=0)}\nonumber\\
    &= \frac{\frac{p(\rvy|\rvz,\rt=1)p(\rvz|\rt=1)}{p(\rvy|\rt=1)}p(\rvy|\rt=0)}{p(\rvz|\rt=0)}\nonumber\\
    &= p(\rvy|\rvz,\rt=1)\underbrace{\frac{p(\rvy|\rt=0)}{p(\rvy|\rt=1)}}_{r(\rvy)}\underbrace{\frac{p(\rvz|\rt=1)}{p(\rvz|\rt=0)}}_{1/Z},\label{eq:balance_p}
\end{align}
where $^*$ uses $p(\rvz|\rvy,\rt=0)=p(\rvz|\rvy,\rt=1)$. The ratio $r(\rvy)$ represents the fraction of the marginal probabilities of $\rvy$, while $Z$ is a normalization constant.

We define the corrected model $q'(\rvy|\rvz)$ following the result from equation~\ref{eq:balance_p}:
\begin{align}
    \hat{q}'(\rvy|\rvz):=\frac{1}{Z}q(\rvy|\rvz)r(\rvy).
\end{align}
The test error for the model $q'(\rvy|\rvz)$ for a representation $\vz$ that has positive support in both train and test ($p(\rvz|\rt=1)>0$,$p(\rvz|\rt=0)>0$) is determined by the train latent test error:
\begin{align}
    \KL(p_{\rvy|\rvz=\vz}^{\rt=0}||q'_{\rvy|\rvz=\vz}) &= \E_{\vy\sim p(\rvy|\rvz=\vz,\rt=0)}\left[\log\frac{p(\rvy=\vy|\rvz=\vz,\rt=0)}{{q}'(\rvy=\vy|\rvz=\vz)}\right]\nonumber\\
    &= \E_{\vy\sim p(\rvy|\rvz=\vz,\rt=0)}\left[\log\frac{p(\rvy=\vy|\rvz=\vz,\rt=0)}{{q}(\rvy=\vy|\rvz=\vz)}\frac{p(\rvy=\vy|\rt=1)}{p(\rvy=\vy|\rt=0)}\frac{p(\rvz=\vz|\rt=0)}{p(\rvz=\vz|\rt=1)}\right]\nonumber\\
    &= \E_{\vy\sim p(\rvy|\rvz=\vz,\rt=0)}\left[\log\frac{\frac{p(\rvy=\vy|\rvz=\vz,\rt=0)p(\rvz=\vz|\rt=0)}{p(\rvy=\vy|\rt=0)}}{\frac{p(\rvy=\vy|\rvz=\vz,\rt=1)p(\rvz=\vz|\rt=1)}{p(\rvy=\vy|\rt=1)}}\frac{p(\rvy=\vy|\rvz=\vz,\rt=1)}{q(\rvy=\vy|\rvz=\vz)}\right]\nonumber\\
    &= \E_{\vy\sim p(\rvy|\rvz=\vz,\rt=0)}\left[\log\frac{p(\rvz=\vz|\rvy=\vy,\rt=0)}{p(\rvz=\vz|\rvy=\vy,\rt=1)}\right] + \KL(p_{\rvy|\rvz}^{\rt=1}||q_{\rvy|\rvz})\nonumber\\
    &=^* \KL(p_{\rvy|\rvz}^{\rt=1}||q_{\rvy|\rvz}),
\end{align}
where $^*$ uses $p(\rvz|\rvy,\rt=0)=p(\rvz|\rvy,\rt=1)$.

\subsection{Relation between the different criteria}
\label{app:crit_relations}
The  different criteria mentioned in section~\ref{sec:criteria} are related by the following expression, which follows from the chain rule of mutual information:
\begin{align*}
    \underbrace{I(\rve;\rvz|\rvy)}_{\text{separation}} + I(\rve;\rvy) = \underbrace{I(\rve;\rvy|\rvz)}_{\text{sufficiency}} + \underbrace{I(\rve;\rvz)}_{\text{independence}}
\end{align*}
From which we can derive the following:
\begin{enumerate}
    \item if $\rvy$ and $\rve$ are independent ($I(\rvy;\rve)=0$) enforcing separation is equivalent to enforcing both sufficiency and independence:
    \begin{align}
        &I(\rve;\rvz|\rvy)+I(\rve;\rvy)=0\nonumber\\
        &\iff I(\rvy;\rve|\rvz)+I(\rve;\rvz)=0
    \end{align}
    
    \item if $\rvy$ and $\rve$ are dependent ($I(\rve;\rvy)>0$), sufficiency and independence are mutually exclusive conditions \citep{Barocas2018}:
    \begin{align}
        I(\rve;\rvy|\rvz)+I(\rve;\rvz)\ge I(\rve;\rvy)
    \end{align}
\end{enumerate}

\section{Datasets}
\label{app:cmnist_probs}

Here we report the conditional probability distributions used to produce the d-CMNIST and y-CMNIST datasets, underlining the commonalities and differences from the original CMNIST distribution.

Across the three datasets:
\begin{itemize}
    \item The first two environments are selected for training.
    \begin{table}[!h]
        \centering
        $p(\rt=t|\re=e)$\\
        \begin{tabular}{l|lll}
         & $e=0$ & $e=1$ & $e=2$ \\ \hline
        $t=1$ & 1   & 1   & 0   \\
        $t=0$ & 0   & 0   & 1  
        \end{tabular}
    \end{table}
    \item The correlation between color $\rc$ and label $\ry$ is positive in the first two environments and negative on the last.
    \begin{table}[!h]
        \centering
        $p(\rc=1|\ry=y,\re=e)$\\
        \begin{tabular}{l|lll}
            & $e=0$ & $e=1$ & $e=2$ \\ \hline
        $y=0$ & 9/10 & 4/5 & 1/10 \\
        $y=1$ & 1/10 & 1/5 & 9/10
        \end{tabular}
    \end{table}
    \item For a given color $\rc$ and digit $\rd$, the corresponding picture is a MNIST digit with label $\rd$ and red ($\rc=1$) or green ($\rc=0$) color. In the discrete settings we assume that pictures contain only color and digit information (ignoring the style) and we construct $\rvx$ as a concatenation of the color $\rc$ and digit $\rd$: $\rvx:=[\rc,\rd]$.
\end{itemize}

\subsection{CMNIST}
\begin{itemize}
    \item All 10 digits $\rd$ the same probability:
    \begin{align*}
        \forall d\in{0,\hdots,9}:\ p(\rd=d)=1/10
    \end{align*}
    
    \item The 3 environments have the same probability to occur:
    \begin{align*}
        \forall e\in{0,1,2}:\ p(\re=e)=1/3
    \end{align*}
    Note that the first two environments will have probability $0.5$ on the selected training set, while the last one will be drawn with probability $1$ from the test distribution.
    
    \item The label $y=1$ is assigned with probability $0.75$ for digits 5-9 and $0.25$ for digits 0-4 across all environments.
    \begin{table}[!h]
        \centering
        $p(\ry=y|\rd=d)$\\
        \begin{tabular}{l|ll}
              & $d<5$ & $d\ge5$ \\ \hline
        $y=0$ & 3/4  & 1/4    \\
        $y=1$ & 1/4  & 3/4  
        \end{tabular}
        \end{table}
\end{itemize}

\subsection{d-CMNIST}
\begin{itemize}
    \item Digits from 0-4 are more likely to occur on the first environment, less likely on the second, while the probability is uniform on the third environment.
    \begin{table}[!h]
        \centering
        $p(\rd\in D|\rve=e)$\\
        \begin{tabular}{l|lll}
                & $e=0$ & $e=1$ & $e=2$ \\ \hline
        $D=[0,4]$   & 3/5   & 1/5   & 1/2   \\
        $D=[5,9]$ & 2/5   & 4/5   & 1/2  
        \end{tabular}
    \end{table}
    
    Digits from 0 to 4 (and from 5 to 9) have the same probability:
    \begin{align*}
        \forall d\in[0,4]:\ p(\rd=d) = \frac{\sum_{d'\in[0,4]}p(\rd=d')}{5}\\
        \forall d\in[5,9]:\ p(\rd=d) = \frac{\sum_{d'\in[5,9]}p(\rd=d')}{5}
    \end{align*}
    
    \item The first environment is more likely than the second:
    \begin{align*}
        p(\re=0) = 1/2\\
        p(\re=1) = 1/6\\
        p(\re=2) = 1/3
    \end{align*}
    Note that this assignments is designed to ensure that marginally the digits $\rd$ have uniform distribution on both train and test.
    
    \item Labels $\ry$ are assigned in the same way as the original CMNIST distribution, based on the digits $\rd$.
\end{itemize}

\subsection{y-CMNIST}
\begin{itemize}

    \item Digits 0-4 are more likely to occur for $\ry=0$ and less likely for $\ry=1$:
    \begin{table}[!h]
        \centering
        $p(\rd\in D|\ry=y)$\\
        \begin{tabular}{l|ll}
              & $y=0$ & $y=1$ \\ \hline
        $D=[0,4]$ & 3/4  & 1/4    \\
        $D=[5,9]$ & 1/4  & 3/4  
        \end{tabular}
        \end{table}
        
    Similarly to d-CMNIST, digits in the same group ([0,4], [5,9]) have equal likelihood:
    \begin{align*}
        \forall d\in[0,4]:\ p(\rd=d) = \frac{\sum_{d'\in[0,4]}p(\rd=d')}{5}\\
        \forall d\in[5,9]:\ p(\rd=d) = \frac{\sum_{d'\in[5,9]}p(\rd=d')}{5}
    \end{align*}
        
    Note that the conditional distribution $p(\rd|\ry)$ is the same as the one of the original CMNIST distribution. 
    
    \item The label $y=0$ is more likely in the first environment and less on the second. The two labels have the same probability on the third environment:
    \begin{table}[!h]
        \centering
        $p(\ry=y|\re=e)$\\
    \begin{tabular}{l|lll}
          & $e=0$ & $e=1$ & $e=2$ \\ \hline
    $y=0$ & 3/5   & 1/5   & 1/2   \\
    $y=1$ & 2/5   & 4/5   & 1/2  
        \end{tabular}
    \end{table}
    
    \item The marginal environment distribution is the same as the one described in the d-CMNIST dataset. This assignment ensures that both labels $\ry$ and digits $\rd$ have marginal uniform distribution on train $\rt=1$ and test $\rt=0$ splits.
\end{itemize}

\subsection{Dataset properties}
Table~\ref{tab:measures} reports some of the mutual information measurements for the three CMNIST, d-CMNIST, and y-CMNIST datasets. Note that only representations that contain digit or no information result in zero concept shift.

\begin{table*}[!h]
\centering
\scriptsize
\begin{tabular}{c@{\hskip 0.1cm}|c@{\hskip 0.1cm}c@{\hskip 0.1cm}c@{\hskip 0.1cm}c@{\hskip 0.1cm}|c@{\hskip 0.1cm}c@{\hskip 0.1cm}c@{\hskip 0.1cm}|c@{\hskip 0.1cm}c@{\hskip 0.1cm}c@{\hskip 0.1cm}c@{\hskip 0.1cm}|c@{\hskip 0.1cm}c@{\hskip 0.1cm}c@{\hskip 0.1cm}}
         & \multicolumn{4}{c|}{Concept shift}                & \multicolumn{3}{c|}{Independence}                   & \multicolumn{4}{c|}{Sufficiency}                                         & \multicolumn{3}{c}{Separation}                            \\ \hline
dataset  & \rotatebox{90}{$I(\ry;\rt|\rvx)$} & \rotatebox{90}{$I(\ry;\rt|\rc)$} & \rotatebox{90}{$I(\ry;\rt|\rd)$}     & \rotatebox{90}{$I(\ry;\rt)$}       & \rotatebox{90}{$I_{t=1}(\re;\rvx)$} & \rotatebox{90}{$I_{t=1}(\re;\rc)$} & \rotatebox{90}{$I_{t=1}(\re;\rd)$} & \rotatebox{90}{$I_{t=1}(\ry;\re)$}       & \rotatebox{90}{$I_{t=1}(\ry;\re|\rx)$} & \rotatebox{90}{$I_{t=1}(\ry;\re|\rc)$} & \rotatebox{90}{$I_{t=1}(\ry;\re|\rd)$} & \rotatebox{90}{$I_{t=1}(\re;\rx|\ry)$} & \rotatebox{90}{$I_{t=1}(\re;\rc|\ry)$} & \rotatebox{90}{$I_{t=1}(\re;\rd|\ry)$} \\ \hline
CMNIST   & 0.219    & 0.283    & \textbf{0.0} & \textbf{0.0} & 0.00143         & \textbf{0.0}    & \textbf{0.0}    & \textbf{0.0} & 0.00854           & 0.00997           & \textbf{0.0}      & 0.00997           & 0.00997           & \textbf{0.0}      \\
d-CMNIST & 0.238    & 0.306    & \textbf{0.0} & \textbf{0.0} & 0.0642          & 0.000636        & 0.0633          & 0.0152       & 0.00588           & 0.0164            & \textbf{0.0}      & 0.0549            & 0.00760           & 0.0481            \\ 
y-CMNIST & 0.238    & 0.306    & \textbf{0.0} & \textbf{0.0} & 0.0331          & 0.0258          & 0.0152          & 0.0633       & 0.0371            & 0.0444            & 0.0481            & 0.00683           & 0.00683           & \textbf{0.0}      \\\hline
\end{tabular}
\caption{Mutual information quantities for the three CMNIST datasets. Note that only a representation containing digit or no information results in zero concept shift. Both color and digit satisfy the independence criterion on CMNIST, while digit information is not satisfying sufficiency or separation for the y-CMNIST and d-CMNIST datasets. }
\label{tab:measures}
\end{table*}

\subsection{Sampling}
Since all the three dataset have uniform digit distribution on both train and test splits ($p(\rd=d|\rt=1)=p(\rd=d|\rt=0)=1/10$), the training dataset are produced using the following sampling procedure:
\begin{enumerate}
    \item Determine $p(\rc,\re,\ry|\rd,\rt=1)$ using Bayes rule:
    \begin{align*}
        p(\rc,\re,\ry|\rd,\rt=1) = \frac{p(\rc,\re,\ry,\rd|\rt=1)}{p(\rd|\rt=1)}.
    \end{align*}
    \item Sample a MNIST picture $\tilde\vx$ with label $d$ from the MNIST dataset.
    
    \item Sample from $p(\rc,\re,\ry|\rd=d,\rt=1)$ to determine color $c$, environment $e$ and label $y$.
    
    \item Color the picture $\tilde\vx$ in green or red depending on the value of $c$ to obtain $\vx$.
    
    \item Return the triple $(\vx,y,e)$
\end{enumerate}
Note that in the reported experiments, step 3) is performed at run-time. Therefore, the same picture $\tilde\rx$ can appear in different environments with different colors if sampled multiple times.

Digits used for the train $\rt=1$ and test $\rt=0$ splits are disjoint, following the traditional train-test MNIST splits.
The code used to generate the three datasets can be found at 
\url{https://github.com/mfederici/dsit}.

\section{Direct Optimization}
\label{app:mi_opt}

In order to optimize for the objective prescribed by the different criteria directly, we define a discrete random variable $\hat\rvx$ obtained by concatenating the random variables representing color $\rc$ and digit $\rd$.
All the extra information in $\rvx$ (e.g. style, position, small rotations, ..) does not interact with the environment $\re$ or label $\ry$, therefore we can safely assert that $\rvx$ is conditionally independent of $\ry$ and $\re$ once $\hat\rvx$ is observed:
\begin{align}
    I(\rvx;\ry\re|\hat\rvx) = 0.
    \label{ass:indep}
\end{align}
At the same time, we can safely assume that color and digit can be identified when a picture $\rvx$ is observed:
\begin{align}
    \exists g\ s.t.\ \hat{\rvx}=g(\rvx) \ \ \implies I(\hat{\rvx};\ry\re|\rvx)=0.
    \label{ass:stat}
\end{align}
Using the assumptions in \ref{ass:indep} and \ref{ass:stat}, we can infer that $\hat{\rvx}$ is a sufficient statistic for $\rvx$. In particular, we can show that any (conditional or not) mutual information measurement involving $\rvx$, $\rvy$ and $\rve$ is unchanged if $\hat\rvx$ is used instead of $\rvx$.\\
For example:
\begin{align*}
    I(\rvy;\rve|\rvx) &= I(\rvx\rvy;\rve)-I(\rvx;\rvy)\\
    &=I(\rvx\rvy;\rve)-I(\rvx\hat{\rvx};\rvy)+\underbrace{I(\hat\rvx;\rvy|\rvx)}_{=0 \text{ using eq.\ref{ass:stat}}}\\
    &= I(\rvx\rvy;\rve)-I(\hat{\rvx};\rvy)-\underbrace{I(\rvx;\rvy|\hat{\rvx})}_{=0 \text{ using eq.\ref{ass:indep}}}\\
    &= I(\rvy;\rve)+I(\rvx;\rve|\rvy)-I(\hat{\rvx};\rvy)\\
    &= I(\rvy;\rve)+I(\rvx\hat\rvx;\rve|\rvy)-\underbrace{I(\hat\rvx;\rve|\rvx\rvy)}_{=0 \text{ using eq.\ref{ass:stat}}}-I(\hat{\rvx};\rvy)\\
    &= I(\rvy;\rve)+I(\hat\rvx;\rve|\rvy)+\underbrace{I(\rvx;\rve|\hat\rvx\rvy)}_{=0 \text{ using eq.\ref{ass:indep}}}-I(\hat{\rvx};\rvy)\\
    &= I(\hat\rvx\rvy;\rve) - I(\hat\rvx;\rvy)\\
    &= I(\rvy;\rve|\hat\rvx).
    \end{align*}
Note that in this proof we also show $I(\rvx;\rvy) = I(\hat\rvx;\rvy)$ and $I(\rvx;\rve|\rvy)=I(\hat\rvx;\rve|\rvy)$.
The proof for the other reported quantity is analogous.

The dashed trajectories in figure~\ref{fig:cmnist_res} are computed by directly optimizing the objectives reported in section~\ref{sec:criteria} using $\hat{\rvx}$ instead of $\rvx$.
We use a $[(10\times2)\times 64]$ normalized matrix $\mW$ to parametrize the encoder $q(\rvz|\rvx)$. Each column of $\mW$ sums to 1 to represent a valid conditional probability distribution.
This matrix maps each combination of digit (10) and color (2) into a representation $\rvz$ consisting of 64 different options using a dot product:
\begin{align}
    \rvz = \mW^T\text{flatten}(\hat\rvx),
\end{align}
in which $\text{flatten}(\cdot)$ represent the operation flattening the $2\times 10$ matrix into a 20-dimensional vector.
Since $\hat\rvx$ is a categorical variable, this encoder can represents any possible function from $[2 \times 10]$ to the latent space of size $[64]$.
We expect the family of encoders parametrized by $\mW$ to be sufficiently flexible since the dimensionality of the representation is bigger when compared to the number of possible inputs.

All mutual information terms in the objective can be computed and differentiated directly, and we can progressively update $\mW$ using stochastic gradient descent until convergence. 
The matrix $\mW$ is initialized randomly and updated using the ADAM optimizer \citep{Kingma2015} with a learning rate of $10^{-3}$.
The optimization stops when the total variation of train and test error is less than $10^{-4}$ for at least 1000 iterations.

The results for different values of the hyper-parameter $\lambda$ are obtained by training a new matrix $\mW$ from scratch for each considered value of $\lambda$.
The parameter $\lambda$ varies from $0$ to $10^6$ for models trained using the independence, sufficiency and separation criteria, while a maximum value of $10$ is used for the information bottleneck experiments since $\lambda=2$ is usually sufficient to obtain a constant representation.
Further details on the procedure can be found at 
\url{https://github.com/mfederici/dsit}.

\section{Training details}
\label{app:nn_train}
Each model in section~\ref{sec:experiments} is composed of the same encoder $q(\rvz|\rvx)$ and decoder $q(\ry|\rvz)$ neural network consisting of multi-layer perceptrons:
\begin{itemize}
    \item Encoder: 
    \begin{center}
        Flatten()$\to$
        Linear(in\_size$=768$, out\_size$=1024$) $\to$\\
        Linear(in\_size$=1024$, out\_size$=128$) $\to$\\
        Linear(in\_size$=128$, out\_size$=64 (\times 2)$)\\
    \end{center}
    with 0.25 dropout probability and ReLU activation in-between each layer.
    Note that a final size of $128=64+64$ is used for the VIB experiments to separately parametrize mean and variance of the stochastic encoder $q(\rvz|\rvx)$.
    \item Classifier: 
    \begin{center}
        Linear(in\_size$=64$, out\_size$=128$) $\to$\\
        Linear(in\_size$=128$, out\_size$=2$)\\
    \end{center}
    with ReLU activation and softmax output normalization.
\end{itemize}
Each model is trained for a total of 25000 iterations with batches of size 256 (with the exception of the IRM and VREx model that use a batch size of 4096) using the ADAM optimizer \citep{Kingma2015} with a learning rate of $10^{-4}$.
The neural networks are trained for a total of 3000 iterations without any regularization, then $\lambda$ is linearly scaled by a constant value after each iteration ( $5\times10^{-3}$ for VREx, $2\times10^{-2}$ for DANN, CDANN and IRM,  $5\times10^{-5}$ for VIB). The increment policy has been chosen to maintain stability while exploring the full range of latent representation for each one of the models.
The train and test cross-entropy error are measures and reported every 2000 training steps across 3 different seeded runs.

The DANN and CDANN models also require an auxiliary discriminator architecture. In both cases we use multi-layer perceptrons: 
\begin{center}
Linear(in\_size$=64 (+2)$, out\_size$=1024$) $\to$\\
Linear(in\_size$=1024$, out\_size$=128$) $\to$\\
Linear(in\_size$=128$, out\_size$=2$)\\
\end{center}
with ReLU activations, softmax output and spectral normalization \citep{Miyato2018} for additional stability. The $+2$ in parenthesis accounts for the extra dimensions that are required for the label concatenation in the CDANN model. In between each update of the encoder and classifier neural network, the discriminator model is trained for a total of 64 iterations with a learning rate of $10^{-4}$. This procedure is required to ensure stability of the adversarial training.
The entirety of the experiments reported in this work required approximately a total of 250 compute hours on Nvidia 1080 GPUs.
The code used to train the architectures is publicly available at \url{https://www.github.com/mfederici/dsit}.

\begin{figure}[!t]
    \centering
    \includegraphics[width=0.9\textwidth]{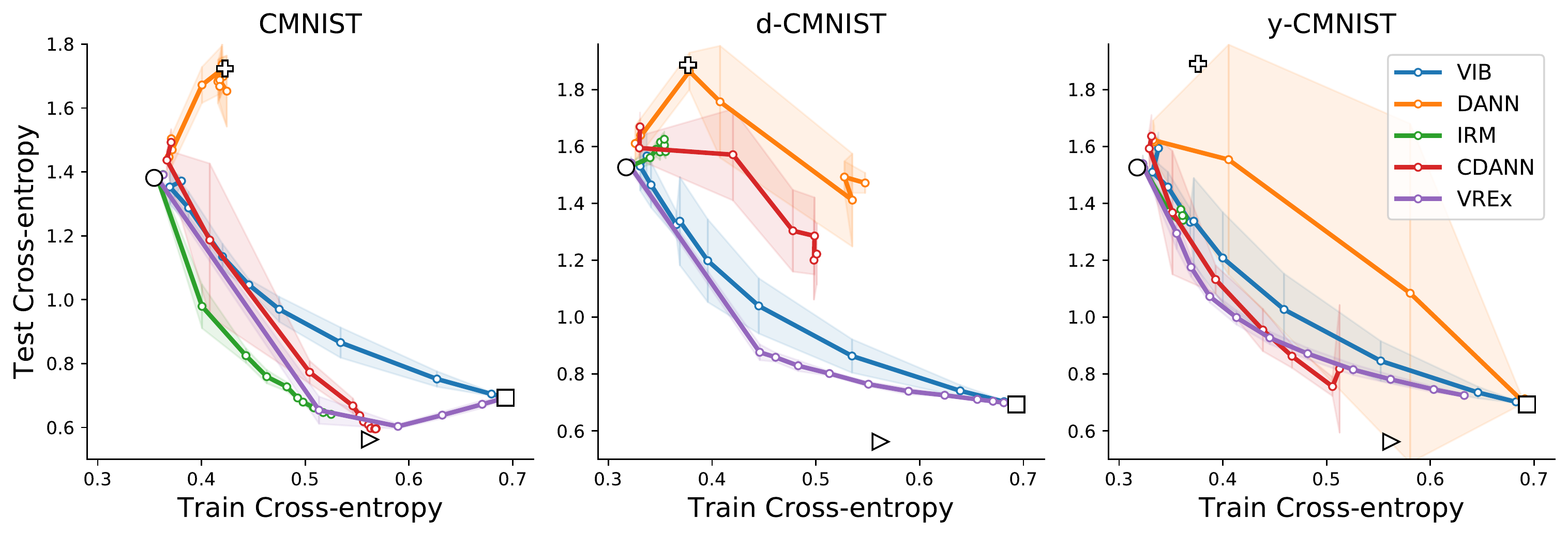}
    \caption{Train and test error on the CMNIST, d-CMNIST and y-CMNIST datasets obtained by training a convlolutional architecture using the same hyper-parameters described in section~\ref{app:nn_train}. The results are consistent with the ones obtained with a multi-layer perceptron (MLP), which are reported in figure~\ref{fig:cmnist_res}.}
    \label{fig:cnn_results}
\end{figure}

\subsection{CNN results}
In order to validate the result obtained with the different regularization strategies, we perform some additional experiments using a convolutional encoder architecture:
\\\\
\begin{center}
Conv(in\_channels$=2$, out\_channels$=32$, kernel\_size$=5$, stride$=2$)$\to$\\
Conv(in\_channels$=32$, out\_channels$=64$, kernel\_size$=5$, stride$=2$)$\to$\\
Conv(in\_channels$=64$, out\_channels$=128$, kernel\_size$=4$, stride$=2$)$\to$\\
Flatten()$\to$ Linear(in\_size$=128$, out\_size$=64$)\\
\end{center}
with ReLU activations and dropout.

The corresponding results reported in Figure~\ref{fig:cnn_results} for the three CMNIST variations are consistent with the ones reported in figure~\ref{fig:cmnist_res}, underlying that the observed behaviour depends on the model objectives more than the chosen architectures.

\section{Hyper-parameters and their effect}
\label{app:empirical_problems}

\subsection{On the number of adversarial steps}

\begin{figure}
    \centering
    \includegraphics[width=\textwidth]{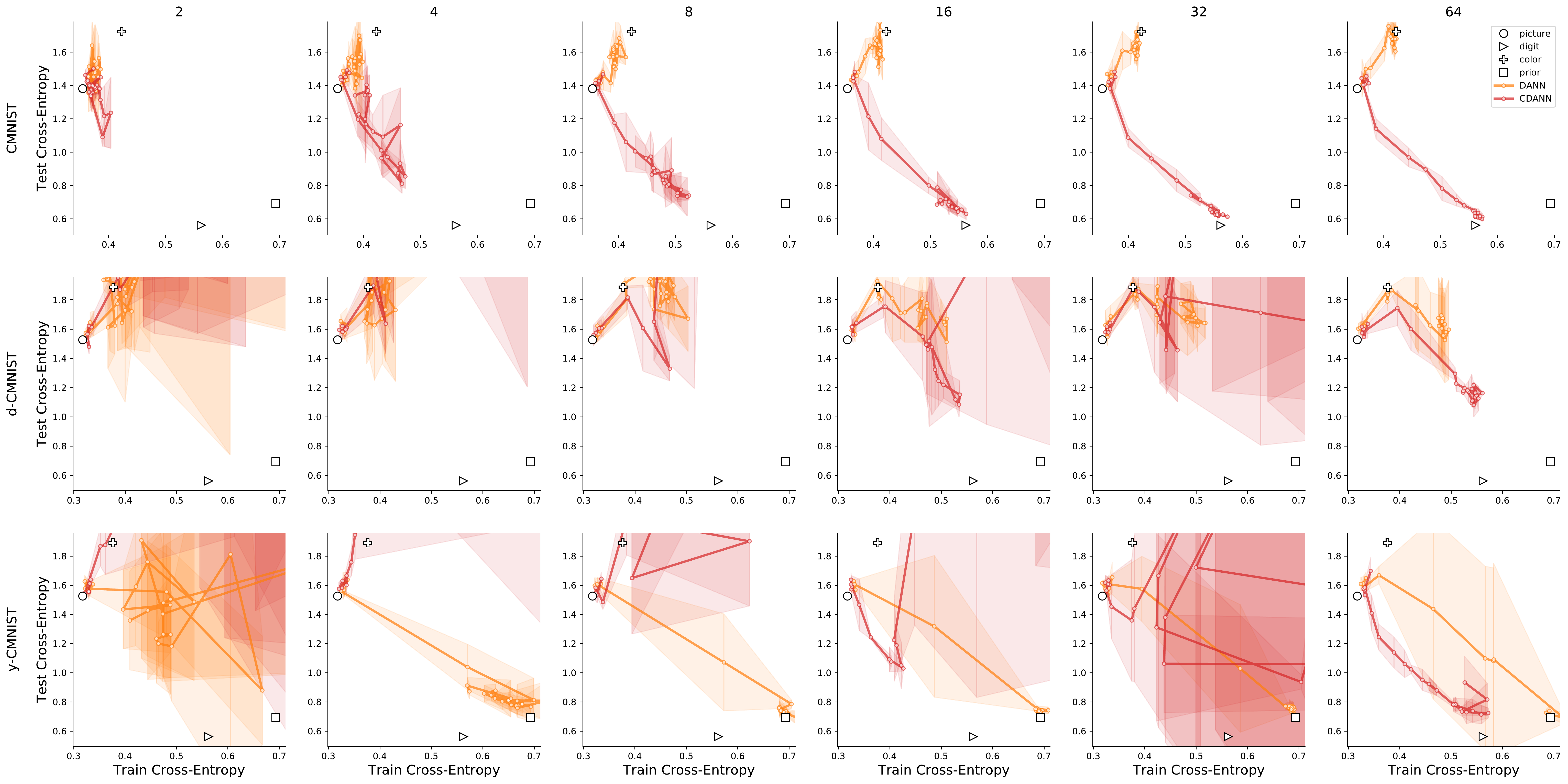}
    \caption{Measurements of train and test cross-entropy for varying number of discriminator training iterations (columns) on the three CMNIST variations (rows) on three different runs. The model convergence and stability is strongly affected by the number of discriminator iterations, especially on the d-CMNIST and y-CMNIST datasets.}
    \label{fig:CMNIST_nadv}
\end{figure}

The results reported in figure~\ref{fig:CMNIST_nadv} clearly show that the effectiveness of the DANN and CDANN model strongly depends on the number of adversarial iterations used for each training step of the encoder model.
The plots in the figure~\ref{fig:CMNIST_nadv} display that an increased number of adversarial iterations helps with both stability and performance of the model. This indicates that the optimality of the adversarial predictors $q(\rve|\rvz)$ and $q(\rve|\rvz,\rvy)$ plays a fundamental role in the training procedure of the model.
\\\\
Note that the training cost and time scales up linearly with the number of adversarial iterations. This becomes especially problematic with bigger models and datasets.

\subsection{On the size of the training batch}
 We observed that both the IRM and VREx models are sensitive to the choice of the size of the training batch used for optimization. We hypothesize that this is due to the fact that both method rely on the computation of a higher order batch statistic (variance and gradient norm).
 
\begin{figure}[!t]
    \vspace{-0.3cm}
    \centering
    \includegraphics[width=\textwidth]{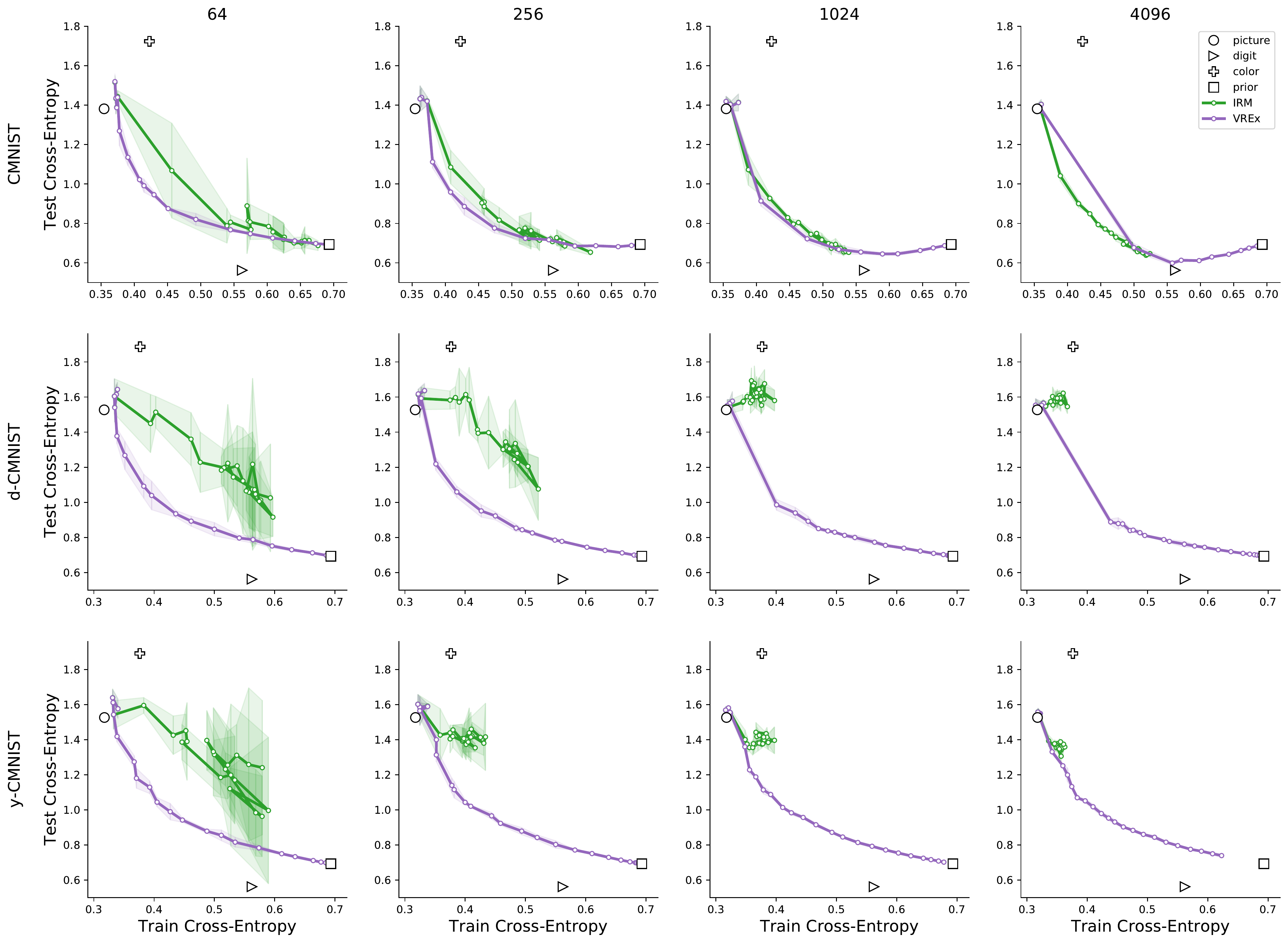}
    \caption{Visualization of the effect of different choices of batch size (different columns) for the IRM and VREx models on the three CMNIST variations (rows) using the MLP architectures described in appendix~\ref{app:nn_train}. Increasing the batch size generally results in better stability and performance at the cost of extra computation time. The models in the pictures are trained for the same amount of iterations.}
    \label{fig:CMNIST_batch}
\end{figure}

 Figure~\ref{fig:CMNIST_batch} reports train and test cross-entropy obtained for different values of batch size.
 We observe that increasing batch size is required to stabilize the training procedure and improve generalization results at the cost of increased memory and computational requirements.


\subsection{Practical limitations of Invariant Risk Minimization}
\label{app:irm_issues}
 We observed that the IRM model is unable to minimize the test error on the d-CMNIST dataset despite the optimality of the sufficiency criterion. We identify two main reasons that can explain this issue:
\begin{enumerate}
    \item The Invariant Risk Minimization introduces a relaxation of the sufficiency criterion:$$\text{Sufficiency} \implies \text{IRM-optimality},\ \ \ \text{IRM-optimality}\ \ \not\!\!\!\!\implies \text{Sufficiency}.$$
    As a consequence, the model can converge to an IRM-optimal solution which is not enforcing sufficiency. Further discussion on the relation between sufficiency and IRM is discussed in \cite{Huszar2019}.
    \item The gradient penalized by the regularization term $\E_{\ve}\vert\vert\nabla\ell_{\ve}(q_{\rvy|\rvz}\circ q_{\rvz|\rvx})\vert\vert^2$ becomes too small to produce meaningful encoder updates.
\end{enumerate}

\begin{figure}[!t]
    \vspace{-0.3cm}
    \centering
    \includegraphics[width=\textwidth]{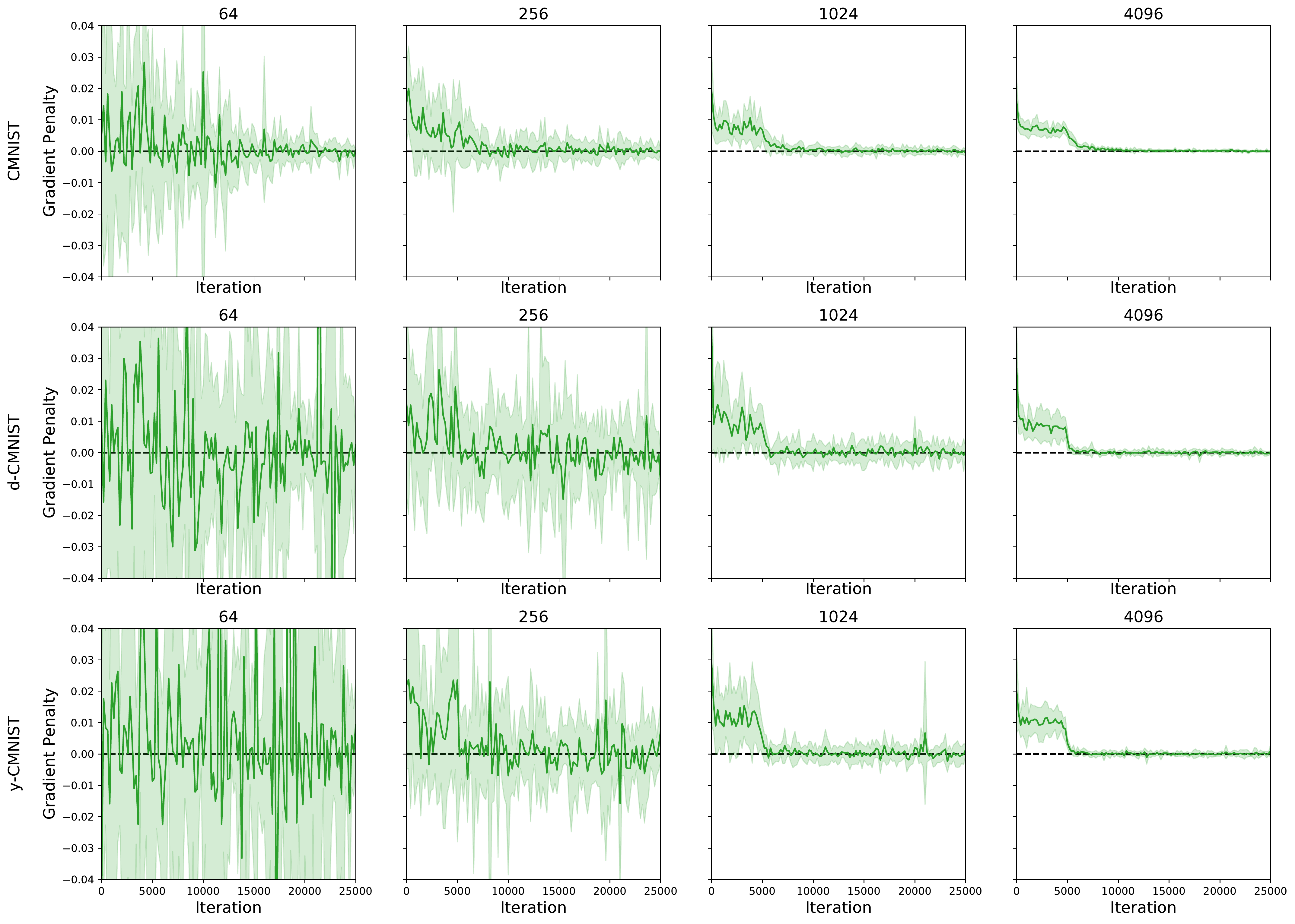}
    \caption{Gradient penalty measured during training of the IRM models using different batch size on the three variations of the CMNIST dataset. Increasing the size of the training batches reduces the variance of the gradient but it does not increase its magnitude.}
    \label{fig:IRM_grad}
\end{figure}

Figure~\ref{fig:IRM_grad} shows that the gradient on the different datasets quickly goes to zero as the regularization strength $\lambda$ is increased. Increasing the batch size results in a decrease in the variance of the gradient estimate, but it does not increase its magnitude. This suggests that either the model already achieved IRM-optimality, or that the gradient is so small that the variance of the regularization term needs to be decreased by orders of magnitudes to result in meaningful updates. Although possible, the latter option would imply that drastically bigger batch size are needed, reducing the applicability of the IRM model to larger settings.


\section{Generalizing constraints from the train distribution}
\label{app:pred_info}

In section~\ref{sec:criteria}, we mention that different criteria implicitly rely on the assumption that enforcing a constraint (or maximizing predictive information) on the training distribution is sufficient to ensure that the constraint holds for the whole distribution, even outside of the selection.
Here we report a simple example in which these assumption are clearly violated to underline the importance of environment variety at training time and pinpoint the limitations of the approaches explored in this work.

Consider the problem of constructing a classifier to distinguish pictures of cows from camels. To simplify our reasoning, we will assume that each picture contains only animal features $\ra\in\{\text{patches, horns, humps, ..}\}$ and background features $\rb\in\{\text{sand, grass}\}$.
The data is collected from different locations $\rve$, which represent the environmental factors.
If the training distribution does not contain any example of cows on sandy terrains or camels on pastures, we can not expect any of the described criteria to successfully classify pictures.
This happens for two main reasons:

\begin{enumerate}
    \item On the training distribution animal and background feature are equally predictive of the label:
    $$I_{\rt=1}(\ry;\ra)=I_{\rt=1}(\ry;\rb)=I_{\rt=1}(\ry;\ra\rb).$$
    As a consequence, we can not expect the encoder $q(\rvz|\rvx)$ to preferentially extract animal features $\ra$, especially when the background $\rb$ is much easier to identify. 
    
    Note that the amount of predictive information for the background $\rb$ on the distribution that has not been selected (camels on pastures, cows on sand for $\rt=0$) is strictly less than the amount of information carried by $\ra$:
    $$I_{\rt=0}(\ry;\rb)<I_{\rt=0}(\ry;\ra)=I_{\rt=0}(\ry;\ra\rb).$$
    This is because not all camels are necessarily in sandy locations, but all of them have humps ($H_{\rt=0}(\ry|\rb)>H(\ry|\ra)=0$).
    In other words, the lack of diversity in the selection makes the background information $\rb$ as predictive as the animal features $\ra$ on the training distribution, although background is not more predictive than the animal features in general.
    
    \item Background information is stable across all the training locations $\rve$:
    $$p(\ry,\rb|\re,\rt=1) = p(\ry,\rb|\rt=1).$$
    This happens since, within the training locations, all cows are on green patches and all camels are on sandy terrains.
    Since the joint distribution of background and label is the same across different training locations, the constraint imposed by both sufficiency and separation criteria cannot remove the background information $\rb$ from the representation $\rvz$.
    Since both $p(\ry|\rb, \rt=1)$ and $p(\rb|\ry,\rt=1)$ appear stable we have:
    $$I_{\rt=1}(\ry;\re|\rb)=0, I_{\rt=1}(\re;\rb|\ry)=0.$$
    In other words, no regularization strength $\lambda$ can force a model trained using the sufficiency or separation (or independence) criterion to remove background information from the representation.
    
    Note that even if background appears to be a stable feature on the selection, it is not stable in general and we have:
    $$I(\ry;\re|\rb)>0, I(\re;\rb|\ry)>0,$$
    since both $p(\ry|\rb)$ and $p(\rb|\ry)$ can change for some environmental conditions that have not been selected. This is because in some (unobserved) locations it is possible to find cows on sandy beaches or camels on grasslands.
    
    
\end{enumerate}
To summarize, lack of diversity in the training environment may compromise the effectiveness of the criteria mentioned in this analysis. This is because models can rely on features that appear to be stable on the training distribution to make predictions. Whenever the stability is due to lack of diversity in the data selection, we cannot expect optimal out-of-distribution performance.

\end{document}